\documentclass{article}

\usepackage[english]{babel}

\usepackage[utf8]{inputenc}
\usepackage{textcomp}
\usepackage{xcolor} 
\usepackage{listings} 
\usepackage[most]{tcolorbox} 
\usepackage{float}
\usepackage{array}
\usepackage{booktabs}
\usepackage{ragged2e}

\usepackage[letterpaper,top=2cm,bottom=2cm,left=3cm,right=3cm,marginparwidth=1.75cm]{geometry}

\usepackage{amsmath}
\usepackage{graphicx}
\usepackage[colorlinks=true, allcolors=blue]{hyperref}

\title{Conversation Routines: A Prompt Engineering Framework for Task-Oriented Dialog Systems}
 
\author{
  Giorgio Robino \\
  \texttt{giorgio.robino@gmail.com}
}

\begin{document}
\maketitle

\begin{abstract}
    \noindent This study introduces \textit{Conversation Routines (CR)}, a structured prompt engineering framework for developing task-oriented dialog systems using Large Language Models (LLMs). 
    While LLMs demonstrate remarkable natural language understanding capabilities, engineering them to reliably execute complex business workflows remains challenging. The proposed CR framework enables the development of \textit{Conversation Agentic Systems (CAS)} through natural language specifications, embedding task-oriented logic within LLM prompts. This approach provides a systematic methodology for designing and implementing complex conversational workflows while maintaining behavioral consistency.
    We demonstrate the framework's effectiveness through two proof-of-concept implementations: a \textit{Train Ticket Booking System} and an \textit{Interactive Troubleshooting Copilot}. These case studies validate CR's capability to encode sophisticated behavioral patterns and decision logic while preserving natural conversational flexibility.
    Results show that CR enables domain experts to design conversational workflows in natural language while leveraging custom functions (\textit{tools}) developed by software engineers, creating an efficient division of responsibilities where developers focus on core API implementation and domain experts handle conversation design.
    While the framework shows promise in accessibility and adaptability, we identify key challenges including computational overhead, non-deterministic behavior, and domain-specific logic optimization. Future research directions include CR evaluation methods based on prompt engineering framework driven by goal-oriented grading criteria, improving scalability for complex multi-agent interactions, enhancing system robustness addressing the identified limitations across diverse business applications.
\end{abstract}

\section{Conversational Agentic Systems (CAS)}

In the LangChain blog article \textit{What is an AI agent?}~\cite{ref1}, Harrison Chase states:

\begin{quote}
\textit{An AI agent is a system that uses an LLM to decide the control flow of an application.}
\end{quote}

This definition resonates with the proposed approach, where business logic is embedded directly within LLM prompts to guide agent behavior. In this paradigm, the prompt itself acts as the primary mechanism for defining control flow.\\

Therefore, in line with Chase’s definition, the conversational agent examples presented here are AI agents: they utilize an LLM to determine the application’s control flow---the train ticket booking or troubleshooting process---based on the embedded CR. This approach contrasts sharply with traditional hard-coded dialog managers, offering greater flexibility and adaptability by defining control flow declaratively through human-readable instructions rather than complex programming code  ~\cite{ref16, ref17, ref18, ref19}.

\subsection{Combining Chat Completion and Function Calling}

A Conversational Agentic System (CAS) can be defined as an AI agent specialized in chat-based interactions, combining the conversational capabilities of LLMs with structured function calls to execute specific tasks and workflows. \\

Unlike basic prompt \textit{text-completion}, which characterizes foundational large language models, a CAS requires an LLM that has undergone specialized fine-tuning on dialogue data. This chat-specific training, exemplified by OpenAI GPT-3.5 models behind applications like ChatGPT, transforms the base LLM’s capabilities into sophisticated dialogue management~\cite{ref2, ref3}. A \textit{chat-completion} model can effectively process the complete sequence of messages exchanged between the user and assistant, along with the system prompt, which is the focus of this study. This dialogue-oriented fine-tuning enables the LLM to maintain conversation coherence, track context across multiple turns, and generate contextually appropriate responses.\\

Using \textit{chat-completion} models, dialogue applications operate through a dynamic feedback loop where user inputs and system responses are appended to a shared \textit{conversation history}. This history, encompassing the system prompt and prior message exchanges, serves as the model’s short-term memory, enabling in-context learning for coherent and contextually aware responses.\\

Effective management of this \textit{conversation history} is vital due to the token limitations of LLMs. While strategies like truncating older messages or summarizing interactions are commonly used to maintain dialogue relevance, these techniques fall outside the scope of this study.

The second foundational prerequisite for enabling LLM-based agents is the use of an LLM pretrained for \textit{function-calling}. The LLM is supplied with a list of available functions (also referred to as tools) that can be implemented in any programming language. This capability to invoke external tools is essential for developing task-oriented applications within specific, closed domains.

\begin{figure}[!ht]
\centering
\includegraphics[width=1\linewidth]{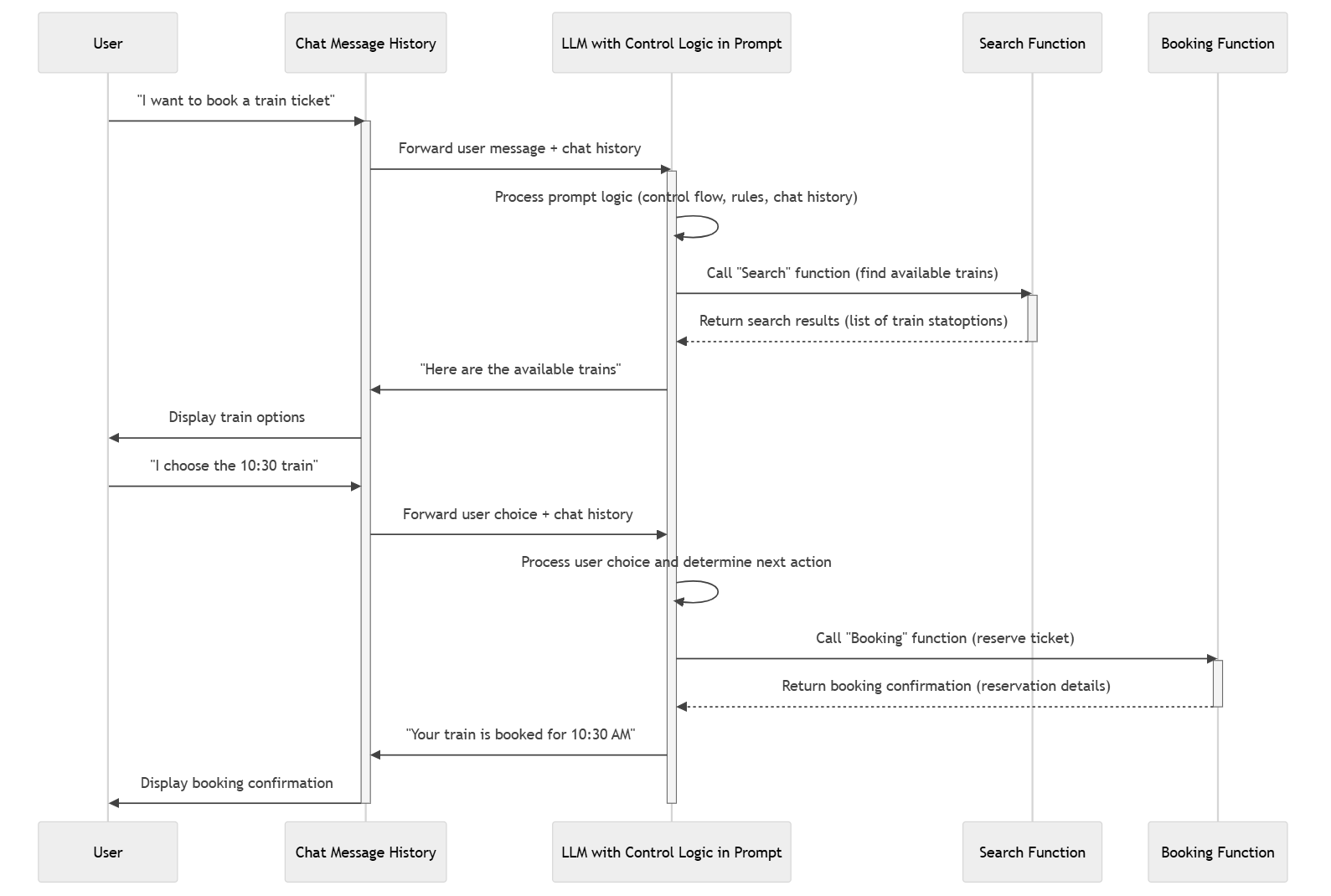}
\caption{\small\label{fig:image_1}A Conversational Agentic System for train ticket booking, showing how the LLM with control logic interacts with external functions.}
\end{figure}

The \textit{function-calling} feature is a critical component that necessitates instruction-tuned LLMs (instruct-models) as a foundational prerequisite for Conversational Agentic Systems. Instruction tuning, as introduced by models such as \textit{InstructGPT}~\cite{ref5}, is pivotal because it equips these models with the ability to bridge natural language and programmatic logic — a capability essential for autonomous agent operations. 
This training enables the models to translate natural language descriptions into structured programming operations, making instruction tuning an indispensable step prior to implementing function calling in agent-based systems. Without this instruction-tuned foundation, models would lack the reasoning capabilities required to accurately interpret user intentions and reliably execute programmatic actions within a conversational framework.

\begin{figure}[H]
\centering
\includegraphics[width=0.8\linewidth]{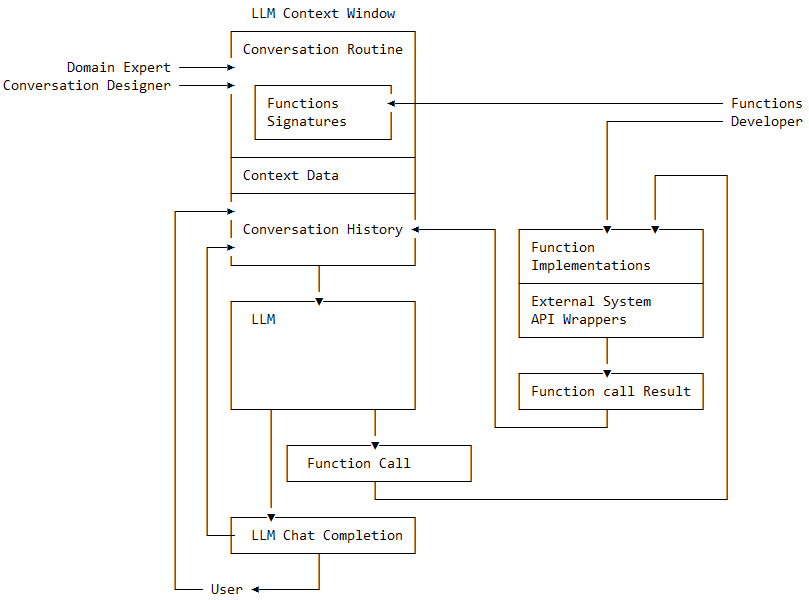}
\caption{\small\label{fig:image_0}A Single-Agent Conversational Agentic System Architecture.
The figure illustrates the interaction between key components: the \textit{Conversation Routine}, contextual data, and conversation history, all encapsulated within the LLM's context window. Inputs from domain experts and conversation designers define the agent's instructions and functions, while developers implement API wrappers and external system calls. 
The system dynamically processes user inputs through the LLM's chat completion capabilities, leveraging modular function calls and seamlessly returning results.}
\end{figure}

\subsection{Multi-Agent CAS Frameworks: Insights from OpenAI SWARM}

A Conversational Agent System (CAS) can be implemented using either a \textit{single-agent} architecture or a \textit{multi-agent} architecture, where multiple agents collaborate to accomplish a task. In omy experiments, I employed a single-agent configuration because our specific use cases did not initially require decomposing the workflow into multiple specialized agents. However, for complex workflows, a multi-agent approach often offers significant advantages in terms of modularity, scalability, and maintainability. \\

The \textit{OpenAI SWARM} framework provides a compelling illustration of multi-agent orchestration in action ~\cite{ref6, ref7}. 
At its core, \textit{SWARM} introduces the concept of \textit{dynamic switching} between specialized agents.  Instead of relying on one monolithic agent to handle every aspect of a conversation or task, \textit{SWARM} empowers distinct agents—each with a narrowly defined role and a tailored system prompt—to take control when their expertise is needed. 
This modular design minimizes unintended interactions that can arise when too many diverse logics are embedded within a single system. \\

\textit{SWARM} also emphasizes \textit{context-aware handoffs}. Each agent operates on a shared conversation history and context variables, allowing it to recognize when a query exceeds its domain of expertise. In such cases, the agent seamlessly transfers control to another agent better equipped to address the request. 
This handoff mechanism ensures that overall coherence is preserved and that the system’s responses remain contextually appropriate throughout the interaction. \\

Another key benefit of the multi-agent approach—as demonstrated by \textit{SWARM}—is the use of \textit{specialized agent routines}. 
Rather than embedding all business logic into one unwieldy prompt, the framework modular routines allow designers to encode conditional logic, iterative processes, and error management directly into each agent’s instructions. This \textit{routines} strategy not only simplifies individual agent design but also makes the entire system more robust and adaptable.
By isolating different parts of the business logic, updates or refinements can be made with minimal impact on the overall system—a critical advantage for scalability and long-term maintenance. \\

In summary, while our experimental work initially employed a single-agent setup for simplicity, the insights derived from the \textit{OpenAI SWARM} framework highlight the significant advantages of adopting multi-agent architectures for managing complex, multi-step workflows. By leveraging \textit{dynamic agent switching}, shared context, and a modular design, multi-agent systems can greatly enhance both the flexibility and efficiency of Conversational Agentic Systems (CAS). This approach not only aligns with contemporary advancements in prompt engineering and LLM-based orchestration but also establishes a scalable foundation for future applications requiring seamless integration of diverse, specialized capabilities. \\

Although the experimental prototypes in this study were implemented using \textit{SWARM}'s single-agent configuration, the concepts and findings presented remain broadly applicable and independent of any specific agent framework. In the final chapter of this paper, I'll explore alternative frameworks and their potential applications in this domain.

\subsection{Embedding Business Logic via Natural Language Instructions}

Traditional deterministic dialogue systems rely on rigid, rule-based state-machine approaches, which often result in inflexible and cumbersome interactions. In contrast, LLM-driven workflows offer greater adaptability by embedding business logic directly within the model’s context through structured natural language prompts. \\

Conversational Agentic Systems are designed to handle conditional logic, iterative processes, and complex procedural sequences, thereby facilitating the execution of moderately complex business tasks. This study explores how embedding business logic within LLM prompts can enhance both functionality and efficiency in closed-domain applications, leading us to ask the following research question:

\begin{tcolorbox}[colback=gray!10, colframe=gray!50, arc=4mm, boxrule=0.5mm, width=\linewidth, center]
\itshape
\textbf{How can natural language instruction prompts be designed to manage procedural logic, conditional flows, and data dependencies so that LLMs can execute multi-step workflows while adhering to specific business constraints?}
\end{tcolorbox}


\section{System Prompts as Conversation Routines}

While CAS excel at natural language interactions and task execution through backend systems, controlling their behavior to enforce business rules remains challenging. This study leverages prompt engineering by embedding procedural logic within \textit{Conversation Routines (CR)}, focusing on system prompts that encode business logic through natural language (\textit{low-code/pseudo-code}) specifications. This enables LLMs to function as workflow interpreters.\\

The meaning of routine was introduced by OpenAI in the context of SWARM, an experimental open-source framework for conversational agents, as described in the cookbook article \textit{Orchestrating Agents: Routines and Handoffs} \cite{ref6}\cite{ref8}. In this article, Ilan Bigio defines a routine as:

\begin{quote}
\textit{The notion of a “routine” is not strictly defined, and instead meant to capture the idea of a set of steps. Concretely, let’s define a routine to be a list of instructions in natural language (which we’ll represent with a system prompt), along with the tools necessary to complete them.[…]}

\textit{The main power of routines is their simplicity and robustness. Notice that these instructions contain conditionals much like a state machine or branching in code. LLMs can actually handle these cases quite robustly for small and medium sized routine, with the added benefit of having “soft” adherence — the LLM can naturally steer the conversation without getting stuck in dead-ends.}
\end{quote}

The Conversation Routines approach outlined in this paper aligns with the \textit{OpenAI SWARM} definition of routines, leveraging structured natural language directives to specify precise agent behaviors.\\

This approach enables conversation designers to develop complex logic without the need for traditional programming languages. By embedding decision-making logic directly into prompts, CR shifts control flow from conventional code to natural language specifications, leveraging the advanced understanding capabilities of LLMs.\\

In this framework, the system prompt serves as a comprehensive software requirements document, providing both procedural instructions and declarative specifications to guide agent behavior effectively.\\

A notable advantage of CR is its accessibility to non-programmers. Conversation designers and domain experts can independently author and refine prompts, reducing reliance on software developers while ensuring flexibility and clarity in agent behavior design. \\

Application development using CR is structured into two primary components with distinct ownership and responsibilities. The first component, managed by conversation designers, consists of writing the system prompt. This prompt incorporates the CR technique and includes function declarations that clearly define the required tools with names, descriptions, parameters, and return values.

The second component, maintained by software developers, is the \textit{Tools} Library — a collection of custom-implemented functions. Each function is thoroughly documented with appropriate docstrings, providing the actual functionality referenced in the function declarations.

\begin{figure}[H]
\centering
\includegraphics[width=1\linewidth]{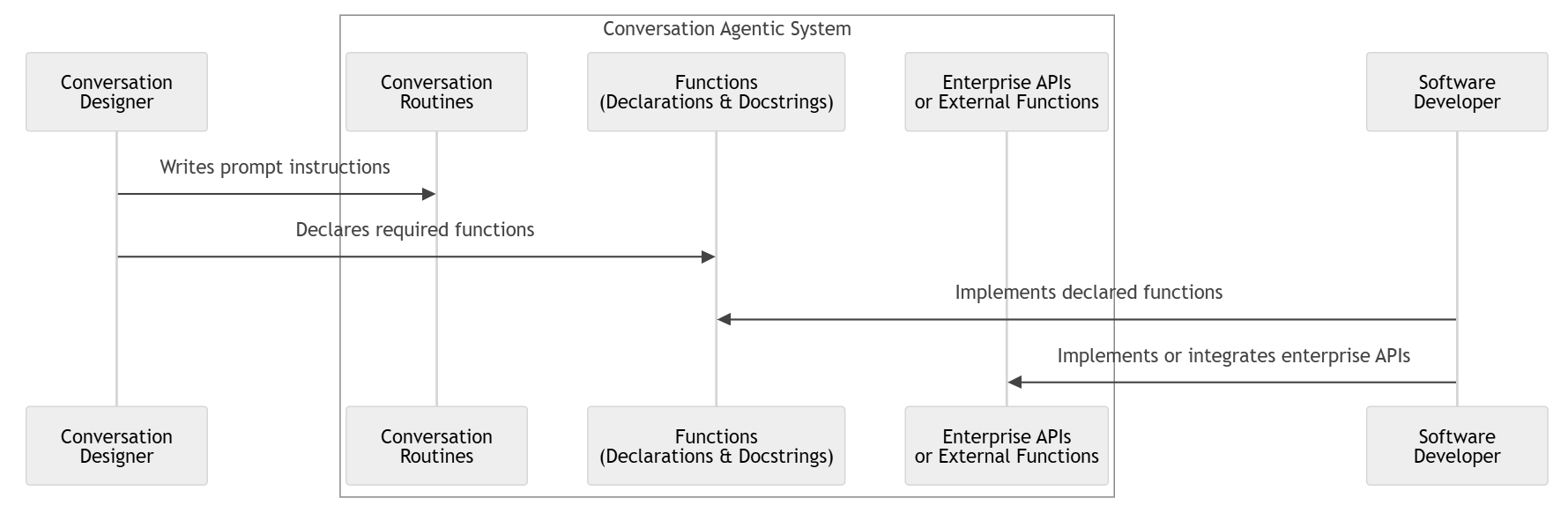}
\caption{\small\label{fig:image_4}Collaboration Between Designers and Developers in a Conversation Agentic System.}
\end{figure}

This clear separation of responsibilities allows conversation designers to concentrate on crafting effective conversational flows and specifying the required capabilities, while developers focus on implementing and maintaining the underlying functional tools. The function declarations act as the interface, ensuring seamless collaboration between the two domains.

\subsection{Common Conversation Routines Components}

The following sections outline the common components of a CR prompt.

\subsubsection{Identities and Purpose}

The foundation of any prompt lies in clearly defining the agent's identity and purpose (\textit{bot-persona}), as well as the user's identity (\textit{user-persona}). This step transcends basic goal-setting by establishing the contextual framework for all subsequent interactions. For instance:

\begin{lstlisting}
You are a specialized assistant focused on helping customers book train tickets.
Primary objectives:
- Guide customers through the booking process
- Ensure accurate information collection
- Maintain natural, helpful interactions
- Handle problems with patience and clarity

User persona examples:
- A first-time traveler unfamiliar with the booking process
- A frequent commuter seeking quick reservations
- A customer managing group bookings with specific preferences
- An international traveler requiring assistance with language or currency
\end{lstlisting}

\subsubsection{Functions Integration Protocol}
A key section focuses on defining how the agent should interact with available tools (external functions). By embedding function specifications also into the system prompt, the agent gains an understanding of both technical capabilities and contextual appropriateness. Example:

\begin{lstlisting}
CORE FUNCTIONS:
- Station Search: Use `search_railway_station()` to search for departure and arrival stations. 
  Support pagination (next, previous, specific page selection) 
  and allow users to refine search queries if needed. 
  Users must explicitly select a station from the available options.
  
- Booking: Use `book_train_ticket()` to finalize the booking. 
  This function requires essential parameters like stations, dates, 
  times, passengers, and class.
  
- Date/Time Handling: Use `get_date_time()` to ensure valid and future dates. 
  If issues arise, provide context-aware suggestions for alternative dates.
\end{lstlisting}
\vspace{1em}
In the above example, when using \texttt{search\_railway\_station()} to find stations, the agent manages paginated results and allows query refinement, integrating responses seamlessly into the conversation. The protocol also emphasizes error management, equipping the agent to recover from common failures and provide user-friendly feedback. This integrated approach ensures the agent remains both technically proficient and contextually aware, enabling efficient and natural interactions.

\subsubsection{Workflow Control Patterns}
The foundation of agent behavior lies in structured workflow patterns that guide decision-making and action sequencing. These patterns serve as essential frameworks for organizing complex interactions, ensuring consistent and reliable agent responses across various scenarios. The following sections explore key workflow patterns - sequential steps with conditionals, iteration logic, and user confirmation protocols - that form the building blocks of robust agent behavior.

\subsubsection{Sequential Steps including conditionals} 
I use systematic indentation patterns to create hierarchical structures for agent behavior specifications. This formatting syntax, similar to Markdown or YAML, enables clear visualization of workflow relationships, with each level of indentation representing a deeper section of behavioral logic. Parent-child relationships between workflow components are expressed through consistent spacing, typically using 2--3 spaces per level, allowing both humans and LLMs to easily parse the control flow and scope of different behavioral instructions.

\begin{lstlisting}
TRAIN_TICKET:

 1. Station Search
    - Get departure station
      * Validate via search
      * Confirm selection
    - Get arrival station
      * Validate via search
      * Confirm selection
 
 2. Journey Details
    - Collect date/time
    - Get passenger count
    - Set travel class
    
 3. Confirmation
    IF all valid:
      - Show summary
      - Process booking
    ELSE:
      - Flag issues
      - Return to input    
\end{lstlisting}

\subsubsection{Iteration Logic} 
A crucial aspect is its ability to specify iteration patterns within workflows. These patterns enable agents to handle repetitive tasks and multi-step processes while maintaining context and progress. When defining iteration logic, designers can embed both explicit loop structures and conditional repetition patterns using natural language constructs. For instance, in moderate complex workflows like train booking, iteration specifications might include both user-driven and system-driven loops:

\begin{lstlisting}
ITERATION CONTROL PATTERNS:

1. User-Driven Iteration
   - Present search results in paginated format
   - While user has not made final selection:
     * Allow navigation between pages
     * Accept selection or refinement request
     * Track current page position
   - Exit loop only on explicit user confirmation

2. System-Driven Validation
   - For each required booking detail:
     * Collect and validate input
     * Maximum 3 validation attempts
     * On failure, offer guided correction
   - Continue until all fields validated or user requests process restart
\end{lstlisting}
\vspace{1em}    
The power of natural language allows for nuanced iteration control, seamlessly integrating loop management with user interaction patterns. This approach to iteration specification proves particularly valuable in scenarios requiring persistent state management and progressive refinement of user inputs, while maintaining the flexibility to break loops based on user satisfaction or system constraints.

\subsubsection{User Confirmation Protocols}
This section establishes critical checkpoints where user confirmation must be obtained before proceeding with the workflow. Rather than relying on assumed understanding, we structure explicit verification points to ensure alignment between user intent and system actions. These verification points can be implemented through natural conversational flows or through deterministic checkpoint functions requiring explicit binary responses. Here’s how we might explain it:

\begin{lstlisting}
Remember to confirm important details naturally, like checking in with
"I've got [this information] so far - is that correct?" before moving forward.
\end{lstlisting}
\vspace{1em}
For critical workflow stages requiring deterministic confirmation:

\begin{lstlisting}
"To proceed with the booking, please review these details: [booking details]

You must confirm with Yes or No.
There are no alternative responses at this stage.
After user response always call user_confirmed_booking(confirmation_flag)"
\end{lstlisting}

\subsubsection{Output Format and Tone}
This section is all about consistency — making sure our agent communicates in a way that’s both professional and engaging. This isn’t about rigid templates, but rather about maintaining a consistent voice and structure throughout the conversation. For example:

\begin{lstlisting}
Keep your responses friendly but focused. 
When presenting options, structure them clearly:

"Here's what I found for your journey:
Morning option: 9:15 AM departure
Afternoon option: 2:30 PM departure
Which time works better for you?"
\end{lstlisting}

\subsubsection{Conversation Examples} 
To further refine agent behavior, interaction examples can be incorporated using \textit{one-shot} or \textit{few-shot prompting} techniques. These examples serve as practical demonstrations of desired conversational flows, tones, and responses, helping to anchor the LLM’s behavior within specific interaction patterns. By including one or more illustrative dialogues, designers can guide the system toward consistent responses in similar contexts.\\

For optimal effectiveness, examples can be tailored to each specific flow or decision point within the workflow. For instance, an example might demonstrate how the agent should confirm customer details during the data collection phase or how it should handle edge cases during the validation phase. These examples act as reference points, ensuring the LLM adheres to predefined interaction standards while maintaining natural and contextually appropriate communication.

\begin{lstlisting}
CONVERSATION EXAMPLE

User: I need a train ticket to London
Agent: I'll help you book a ticket to London. First, I need some details:
1. What's your departure city?
2. What date would you like to travel?
3. Do you prefer morning or evening travel?

Let's start with your departure city.
\end{lstlisting}
\vspace{1em}
This structured approach creates a natural flow from understanding the basic purpose through to handling specific interactions, all while maintaining a conversational tone that makes the agent feel more approachable and helpful. Each section builds upon the previous ones, creating a comprehensive but flexible framework for our conversational agent.\\

The sections listed above represent the basic common components of the framework. In the next chapter, I will present detailed real-world use cases as examples.\\

\section{Conversation Routines in Practical Use Cases}

In my previous white paper, \emph{“A Conversational Agent with a Single Prompt?”}~\cite{ref9}, I presented three experimental use cases that demonstrate how Large Language Models can be programmed through structured prompts to achieve sophisticated task-oriented dialogs.\\

\textit{Virtual Caregiver for Telemedicine}: This case study involved a virtual caregiver designed for remote patient monitoring. This agent conducted daily check-ins, collecting vital health metrics such as temperature and blood oxygenation while maintaining empathetic, natural conversations. The system demonstrated a key strength of prompt-engineered LLMs: the ability to accommodate natural conversational digressions while reliably returning to structured data collection tasks.\\

\textit{Customer Care Assistant}: This case study featured a customer service agent that bridged the gap between free-form user descriptions and structured ticketing systems. This implementation highlighted dual advantages of natural language programming: simplifying development complexity while enabling natural human-like interactions. Users could express their issues conversationally, while the agent systematically gathered the stru
ctured data required for backend processing.\\

\textit{Virtual Job Interviewer}: This prototype showcased a virtual recruiter that dynamically interviewed candidates by analyzing their CVs against job descriptions embedded in the prompt. This system demonstrated sophisticated cognitive abilities, probing candidates’ strengths, weaknesses, and potential mismatches while maintaining a structured yet adaptive conversational flow. The implementation illustrated how LLMs can perform complex workflows within defined conversational settings.\\

During the development of these prototypes, I utilized the OpenAI GPT-3 (\texttt{text-davinci-003}) \textit{text-completion} model, preceding the release of the GPT-3.5 (\texttt{gpt-3.5-turbo)} \textit{chat-completion} model.\\

These prototypes demonstrated the LLM’s remarkable capacity for fluid dialogue while staying aligned with assigned objectives. The agents exhibited an extraordinary ability to identify the conclusion of conversations when the programmed goal was achieved, along with maintaining an internal state of completed steps and tracking state variables effectively.\\

Building on these previous experiments, I now present two new use cases where CR was applied in a more systematic manner. Each use case includes a description of the application problem, the specific prompts used, and illustrative conversation excerpts.

\begin{enumerate}
    \item \textbf{\textit{Train Ticket Booking System}}: This use case demonstrates how embedded business logic, expressed through CR, enables a Conversational AI System (CAS) to guide users through the complexities of purchasing train tickets.

    \item \textbf{\textit{Interactive Troubleshooting Copilot}}: In this example, a CAS assists technicians in diagnosing and troubleshooting machinery issues. The agent guides users through structured diagnostic procedures and provides access to relevant information, thereby streamlining factory-floor operations.
\end{enumerate}

For these two applications, I employed the OpenAI \texttt{GPT-4o-mini} model as the reference LLM, selected for its extensive 128k-token context window, function-calling capabilities, and low latency~\cite{ref10}. Each use case was implemented as an agent using the OpenAI \texttt{SWARM} framework~\cite{ref7, ref8}. While the multi-agent architecture was used in this study, it is not a mandatory choice, and a detailed exploration of multi-agent systems is beyond the scope of this work.\\

Since my native language is Italian, I conducted interactions in both English and Italian, with all conversation excerpts presented below in Italian. However, I retained English as the preferred language for prompt writing, as previously observed in my experiments~\cite{ref11}. 
Notably, the LLM used demonstrates strong multilingual capabilities, effectively handling multiple natural languages, including Italian, even with minimal language-specific instructions.

\begin{lstlisting}
LANGUAGE:
- Communicate in Italian
- Use second person form for instructions
\end{lstlisting}


\section{Use Case 1: Train Ticket Booking System}

The first use case presents a demonstrative scenario in which a CAS facilitates the purchase of a train ticket. The agent is integrated with backend functions, enabling it to retrieve the current time, query a railway station names database, identify available train connections between specified origin and destination stations, and generate a ticket in a predefined format. A key challenge lies in constructing a prompt that encapsulates the complete business logic necessary for an efficient booking workflow. This process generally involves acquiring and validating user-provided information, followed by invoking a backend function to finalize the booking and payment. \\

Traditional implementations of such workflows rely on hard-coded scripts developed using dialogue management frameworks~\cite{ref16, ref17, ref18, ref19}. This approach necessitates anticipating all potential user actions, including digressions, out-of-scope requests, and error handling. As a result, script complexity increases rapidly, leading to substantial development overhead, increased costs, and a rigid, unnatural user experience. In contrast, the approach explored in this article utilizes a CAS guided by a comprehensive prompt encoding business logic via CR. Rather than hard-coding every possible interaction scenario, the LLM autonomously manages user interactions while adhering to the guidelines and rules specified within the instructions. \\

The CR approach redefines the role of the conversation designer, expanding their responsibilities to include not only traditional interaction design specifications but also the encoding of business logic. As prompt designers, they now craft system behavior, tone of voice, user and assistant personas, and best practices for conversational interaction design - all articulated through natural language specifications within the prompt. \\

A condensed and illustrative summary of the prompt employed in this use case is provided below; however, the complete prompt is more detailed and structured, with certain specifics omitted for brevity. For full details on the train ticket booking functionalities—including the complete workflow description, function signatures, and docstrings that specify behavior, required variables, and the precise data format for function arguments provided by the LLM—please refer to Appendix~\ref{appendix:trainfunctions}.\\

Although the prompt instruction set is expressed in natural language, it adheres to a structured format to facilitate LLM comprehension. This format explicitly defines the logic and principles governing LLM behavior, such as the step-by-step workflow outlined in the \texttt{INTERACTION WORKFLOW} section, the data validation rules specified in the \texttt{INFORMATION COLLECTION} section, and the error handling procedures detailed in the \texttt{ERROR HANDLING} section. \\

\begin{lstlisting}[breakatwhitespace=false]
OBJECTIVE:
Assist users in booking train tickets through clear, user-friendly interactions.

CORE FUNCTIONS:
- Station Search: Use `search_railway_station()` 
  to search for departure and arrival stations. 
  Support pagination (next, previous, specific page selection) 
  and allow users to refine search queries if needed. 
  Users must explicitly select a station from the available options.
- Booking: Use `book_train_ticket()` to finalize the booking. 
  This function requires essential parameters like stations, dates, 
  times, passengers, and class.
- Date/Time Handling: Use `get_date_time()` to ensure valid and future dates. 
  If issues arise, provide context-aware suggestions for alternative dates.

INFORMATION COLLECTION:
- Departure and Destination: Stations are selected through the `search_railway_station()` function. 
  If multiple pages of results exist, navigate using pagination 
  and present results clearly to the user.
- Date and Time: Collect the departure and return dates in 'YYYY-MM-DD' format. 
  Times can be flexible inputs like "early morning" or exact times like "08:30".
- Passenger Details: Collect the number of passengers (positive integer) 
  and the travel class (strictly '1st' or '2nd').

INTERACTION WORKFLOW:
1. Search Phase: 
   - Use `search_railway_station()` for departure and arrival stations.
   - Handle paginated results (e.g., "Page 1 of 3") 
     and offer options to navigate, refine, or select a station.
2. Validation Phase: 
   - Once all required information is collected (stations, dates, times, passengers, class), 
     present a full summary of the booking details.
   - Explicitly ask the user to confirm with "YES" or "NO".
3. Booking Phase: 
   - If the user confirms, call `book_train_ticket()` with the gathered details.
   - If the function returns an error, inform the user, suggest retry options,
     or guide them to contact support.
   - If successful, display all the booking receipt details clearly to the user.

ERROR HANDLING:
- If the booking fails, provide the user with a clear, polite message.
- Allow the user to revise incorrect information at any point during the process.
- If multiple failures occur, offer the user a chance to restart the process 
  or save partial data for later use.

USER GUIDANCE AND BEST PRACTICES:
- Prioritize clear, explicit communication. 
  Provide simple navigation options (next, previous, refine) when browsing station results.
- Ensure flexibility for date/time input (accepts "morning", "08:30", etc.) 
  and strict validation for critical fields (dates, class, and passenger count).
- Ask for explicit confirmation before making any booking to prevent user errors.
- Gracefully handle booking errors and always offer corrective options to users.
\end{lstlisting}
\vspace{1em}

The prompt also incorporates a catalog of available back-end functions (APIs) that the agent is authorized to invoke. These functions, such as \texttt{search\_railway\_station()}, \texttt{book\_train\_ticket()}, and \texttt{get\_date\_time()}, are referenced within the prompt to execute specific tasks.\\

These external functions are typically developed by a separate technical team, establishing a clear demarcation of responsibilities between prompt designers ($\sim$no-coders) and function developers ($\sim$coders).

\begin{figure}[H]
\centering
\includegraphics[width=0.8\linewidth]{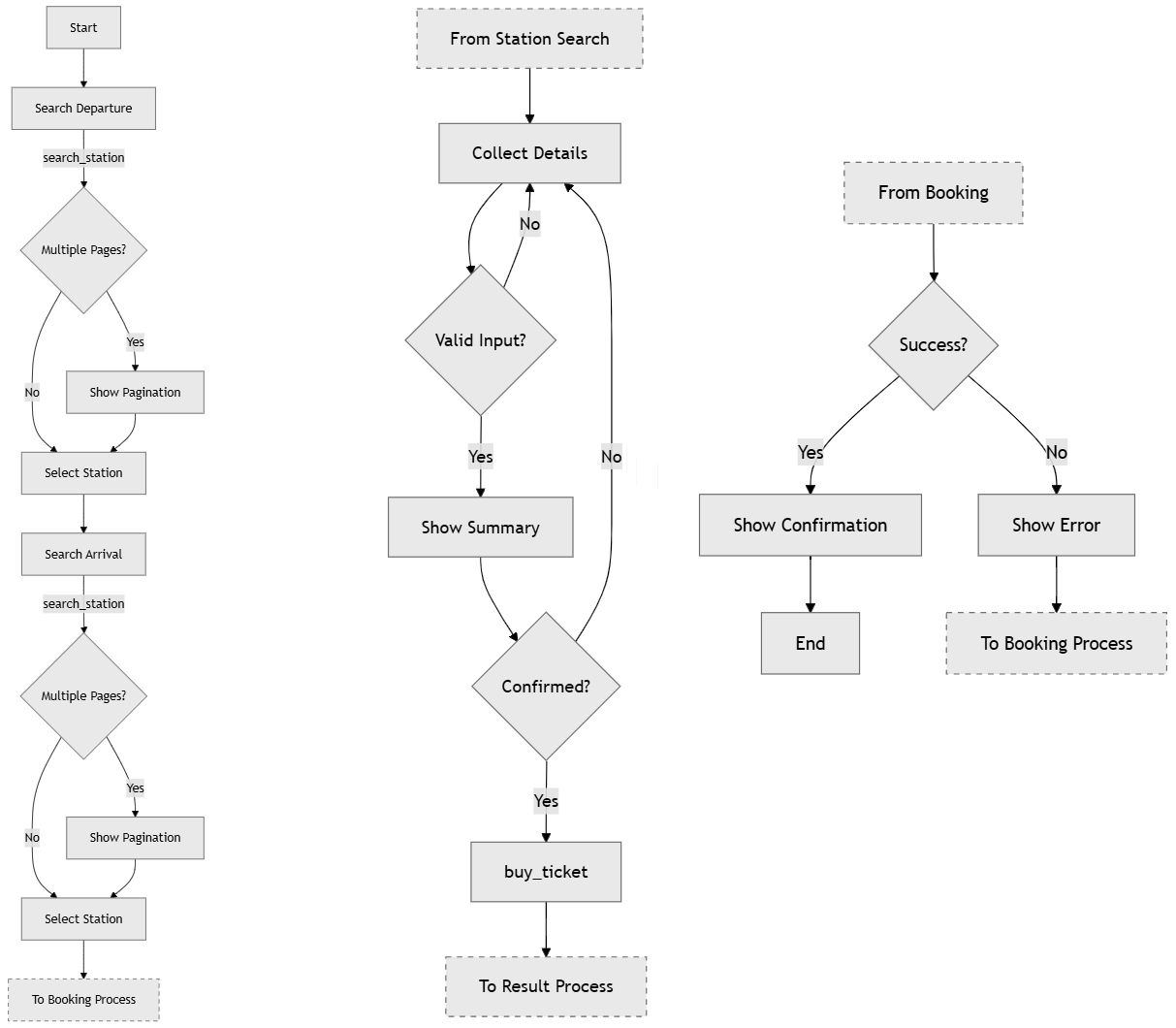}
\caption{\small\label{fig:image_2}Flowchart visualization of the train ticket booking workflow as specified in the LLM prompt. The diagram was generated by prompting an LLM to create a Mermaid flowchart representation of the prompt.}
\end{figure}

\subsection{A Digression on Best Practices for Tools Design}

Although the primary focus of this paper is on Conversation Routines as a prompt engineering technique for crafting system prompts, a significant portion of the CAS development effort is dedicated to defining effective (e.g., Python) functions to provide backend services (e.g., booking services). It is important to highlight that these backend functions are integral to the problem domain and would be required regardless of whether the workflow was implemented using a deterministic, hard-coded software tool.\\

A critical consideration is the design of functions that can be seamlessly invoked by an LLM. A crucial aspect of this design is the management of function state persistence. Ideally, functions should be stateless, adhering to the principles of functional programming, with the LLM maintaining state by providing the appropriate arguments for each function invocation.\\

Conversely, when implementing functions with stateful logic, additional precautions are necessary. In such cases, the state must be tracked and made accessible to the LLM. This can be achieved through shared memory mechanisms or by embedding the relevant state information within the LLM’s prompt context, ensuring continuity across multiple function invocations.\\

To illustrate this point, challenges were encountered when implementing a function to search for railway stations within a dataset of thousands of entries, enabling users to navigate the results using pagination and standard navigation commands (e.g., next page, previous page, first page, last page). Initially, the implementation employed a Python class to manage the state of the instantiated class, necessitating the maintenance and explicit sharing of a transaction ID, thereby complicating the LLM-function interface.\\

A more straightforward solution, however, involved designing a stateless function where the page number is passed as an argument. With this approach, the LLM maintains the current page number internally and provides the appropriate page number based on user navigation requests expressed in natural language with implicit references (e.g., “I don’t see the station I’m looking for”, when on page 1 of 10).\\

In this scenario, the LLM correctly requested page 2 via the function call:

\begin{lstlisting}
search_railway_station(query: "...", page=2)
\end{lstlisting}

This demonstrates the LLM’s ability to handle state transitions without requiring explict state management, such as storing the page variable in the LLM’s context window. The function’s docstring is provided below:

\begin{lstlisting}
def search_railway_station(query: str, page: int = 1) -> str    
    """
    Perform a search for railway station names based on a query 
    and display results in a paginated format.
    
    Args:
        query (str): The search query, consisting of one or more space-separated words.
        page (int, optional): The current page number to display (1-based indexing). 
        Defaults to 1.
    Returns:
        str: A formatted string displaying the search results for the specified page, 
        including pagination details.
    Behavior:
        - Matches are case-insensitive.
        - If no matches are found, a message will indicate zero results.
        - Automatically handles pagination and displays the total number of pages and results.
    Example:
        >>> results = search_railway_station("Genova", page=1)
        >>> print(results)
        Found 2 total results (Page 1 of 1):
        1. Genova Principe
        2. Genova Brignole
    """
\end{lstlisting}

gg
\subsection{Experimental Results: Analysis of a Train Booking Dialog Session}

Let’s analyze a dialog session example where the user’s primary intent is to book a train ticket from Genova to Roma for the following morning, with preferences for traveling alone, first-class seating, and an early departure around 6:00. The \textit{Train Ticket Booking System} dialog session excerpt, provided in Appendix~\ref{appendix:traindialog}, exemplifies the agent’s adherence to prompt instructions for managing complex conversations, particularly in handling station searches and confirming travel details.\\

The agent effectively manages the interaction by adhering to a predefined logical flow while dynamically adapting to the user’s input. It systematically gathers all necessary details, handles ambiguities—such as unclear station names—through paginated search results, and maintains transparency by confirming decisions and summarizing key details throughout the process.
The interaction begins with the agent requesting essential information, such as the departure station, number of passengers, and travel class. When the user is unsure of the station name and provides ambiguous input (\textit{“e’ una staziuone di Genova ma non mi ricordo il nome…”}), the agent does not halt but instead offers to refine the search by providing additional pages of results. Once the user identifies the correct station (\textit{“Nervi”}), the agent seamlessly resumes the flow, confirming the details and proceeding to gather the preferred departure time. This demonstrates the agent’s ability to maintain a user-centered dialog approach by patiently guiding the user through structured queries. \\

The agent correctly leverages available external functions, such as \texttt{search\_railway\_station()}, to retrieve station information, showcasing its ability to interact with external systems as instructed in the prompt. By dynamically responding to user input, the agent clarifies preferences, such as the travel class and departure time, ensuring that no critical details are overlooked. Transparency is maintained throughout the interaction by confirming decisions at key stages and summarizing the collected details before requesting explicit user confirmation. Once all the necessary information is finalized, the agent executes the booking via the \texttt{book\_train\_ticket()} function, providing a comprehensive summary of the travel itinerary, seat details, and total cost. \\

This iterative process highlights the agent's application of behavior rules defined in the prompt, particularly its adherence to requiring user confirmation before executing decisive actions. The session demonstrates the agent's ability to combine structured dialog management, robust error handling, and seamless integration with external systems, resulting in an efficient and user-friendly ticket booking experience.

\begin{figure}[H]
\centering
\includegraphics[width=0.9\linewidth]{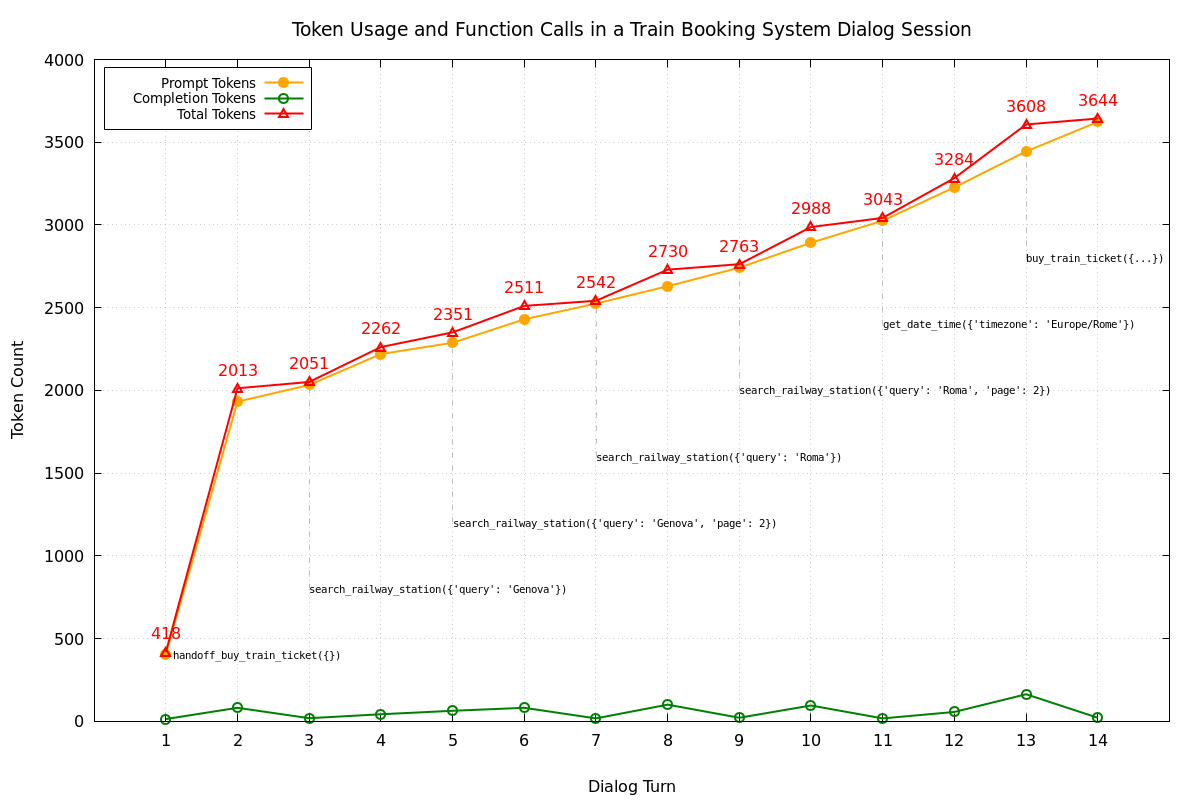}
\caption{\small\label{train_booking_tokens.png}Token consumption increases linearly with context window accumulation, as shown in a 14-agent-turn dialog session using 36,208 tokens over 3 minutes and 41 seconds. Functions were correctly called, with \texttt{search\_railway\_station()} invoked 4 times to resolve source and destination ambiguities. The token count grows from an initial 2013 to 3644 as conversation history and function responses are progressively added to the context window.}
\end{figure}

In conclusion, the dialog interaction highlights the agent’s capability to manage complex conversations efficiently. It dynamically handles ambiguous input, maintains state across multiple turns, and adheres to a predefined logical flow. These qualities result in an effective and seamless user experience, demonstrating the value of using Conversation Routines to structure task-oriented dialog systems.\\

The Conversation Routine prompt consistently delivered excellent user experience results. Across more than 50 test runs, the system successfully booked a train that met the specified requirements in nearly all cases. Challenges arose only when I intentionally introduced non-collaborative interactions, such as requesting a nonexistent station name. Even in these cases, the booking process reliably adhered to the confirmation step, ensuring that the final purchase was completed with accurate data, though it sometimes required additional turns to resolve ambiguities. Conversely, when all necessary data was provided collaboratively, the LLM consistently “converged” on a successful booking within just a few turns, demonstrating its efficiency in handling straightforward interactions.\\

A minor issue I experienced was the LLM’s tendency to output text in markdown format, even when the prompt explicitly requested avoiding special text formatting. A possible solution could involve fine-tuning the general-purpose LLM to precisely follow the desired syntax (e.g., avoiding special formatting). This aspect is potentially significant for voice-interface solutions. However, we consider this a minor problem since it is solvable via fine-tuning or post-processing of the generated text, which lies outside the primary scope of this study.

\newpage
\section{Use Case 2: Interactive Troubleshooting Copilot}

As a proof of concept demonstrating the capability of a LLM to facilitate real-time, task-oriented dialogues, the Interactive Troubleshooting Copilot was developed as a more complex application. This conversational assistant is designed to support technicians in troubleshooting and repairing industrial systems through continuous interactive guidance while adhering to predefined business logic embedded within the system prompt~\cite{ref12}.\\

The proposed system is particularly beneficial for technicians with limited troubleshooting expertise and unfamiliarity with the equipment involved. It provides a conversational interface through which the user can describe equipment status observations and detail the actions performed during the troubleshooting process. Furthermore, the assistant can address ancillary inquiries regarding equipment specifications and details while maintaining focus on the repair task, guiding the operator back to the appropriate step in the procedure as necessary.\\

The assistant interacts with the user via voice and provides step-by-step guidance through the troubleshooting procedure. The process commences with an initial issue description and proceeds by retrieving relevant troubleshooting documentation, employing a \textit{Retrieval-Augmented Generation (RAG)} approach. This methodology enables the assistant to offer continuous, contextually relevant guidance throughout the repair process, ensuring operator adherence to the documented procedure with precision.

\subsection{The Troubleshooting Procedure Manual as Workflow Guideline}

A fundamental challenge in leveraging a LLM for task-oriented dialogues is conveying the specific business logic or processes required to complete a task, such as a repair procedure. The central concept is to utilize a human-readable document outlining the procedure, ideally one already available to technicians.\\

For this study, a \textit{Conveyor Belt Troubleshooting Procedure Manual} was developed — a comprehensive document designed to guide technicians through the diagnosis and repair of industrial conveyor belt systems. This document provides detailed, step-by-step instructions, each associated with specific component codes (e.g., \texttt{QDE-9900-PRO},\texttt{ RLT-8450-V2}), ensuring accurate identification of the components involved. The complete document is available in Appendix~\ref{appendix:troubleshootingprocedure}. \\

The procedure incorporates conditional logic and decision points, where the subsequent step is contingent upon the technician’s observations (e.g., detecting misalignment, damaged parts, or sensor malfunctions). Additionally, the procedure includes iterative loops, requiring the technician to revisit previous steps if anomalies persist following an initial resolution attempt. Visualization of the aforementioned procedure as a flowchart reveals a process consisting of 18 discrete steps. While the logic ranges from relatively simple to moderately complex, the procedure incorporates a significant number of conditional branching points and iterative loops, ensuring a systematic and repeatable troubleshooting process.\\

The manual provides a systematic approach for diagnosing and resolving issues in an industrial conveyor belt system. It covers mechanical blockages, belt misalignment, roller wear, sensor malfunctions, and electrical faults. Steps include visual inspections, component-specific checks (e.g., tension rollers, sensors), emergency protocols, repairs, and final testing. Decision points guide technicians through conditional paths (e.g., "\textit{if objects are stuck, go to Step 3; if misaligned, go to Step 7}"). Key components are identified with codes (e.g., \texttt{CNV-NT2024-A} for the belt) for precise maintenance.

\begin{figure}[H]
\centering
\includegraphics[width=1\linewidth]{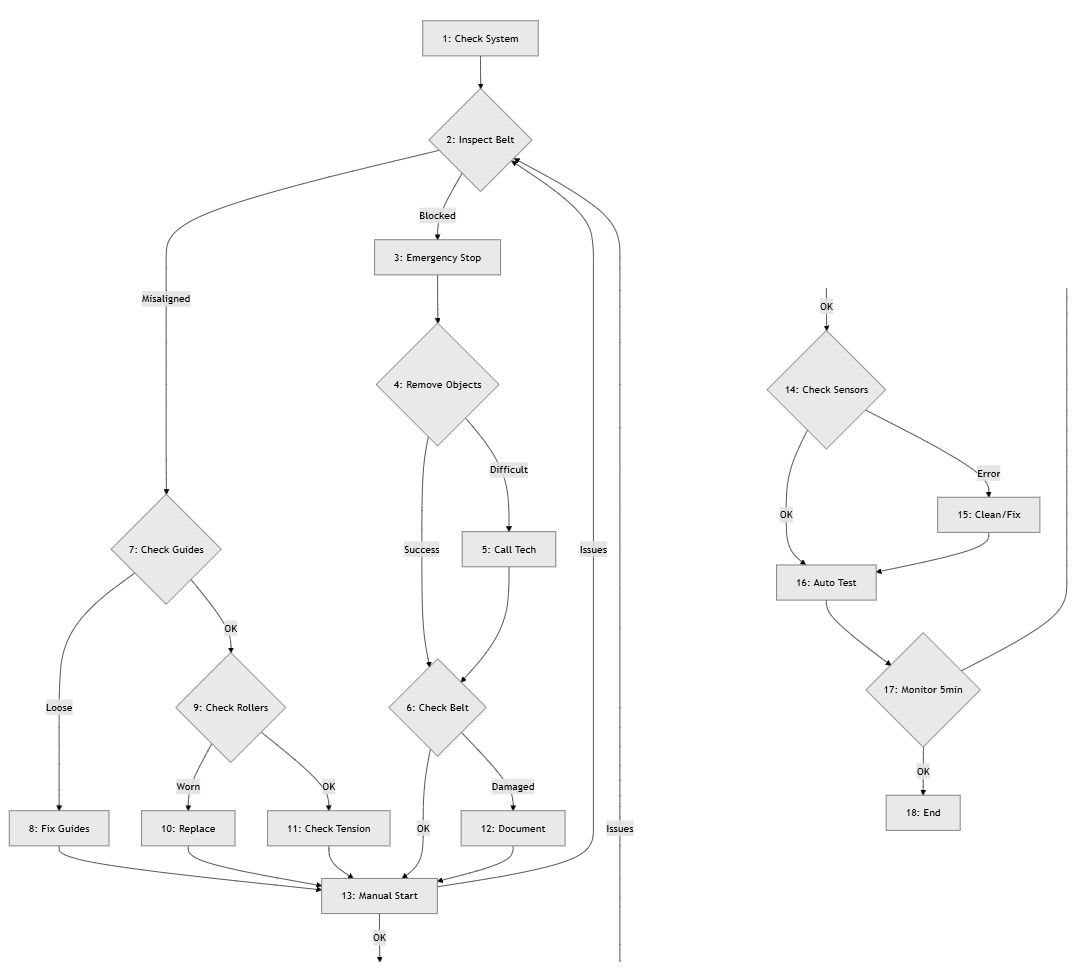}
\caption{\small{\label{fig:image_3}Simplified flowchart representation of the troubleshooting procedure described in the Conveyor Belt Troubleshooting Procedure Manual.
The diagram was generated using an LLM prompt to create a Mermaid-based flowchart.}}
\end{figure}

\subsection{Embedding Troubleshooting Manual Business Logic in the Prompt}

The CAS developed for the use case is designed to assist technicians in diagnosing and repairing conveyor systems. It defines the assistant’s role as an expert, with core functions including retrieving troubleshooting steps, providing component details, and generating reports. \\

The system maintains an internal state, tracking the current step, historical progress, safety status, and component information to ensure both procedural continuity and adherence to safety protocols.\\

The troubleshooting process is structured into four distinct stages and is orchestrated by a multi-agent architecture consisting of two agents:

1. a \textit{\textbf{Troubleshooting Assistant Agent}} manages the \textit{Initial Assessment} stage by retrieving the relevant instructions using a \textit{Retrieval-Augmented Generation (RAG)} approach and verifying the necessary safety protocols. During the \textit{Step-by-Step Execution} phase, this agent guides the technician through each sequential repair step, ensuring compliance with all safety measures. If additional information, such as component specifications, is required, the \textit{Part Inquiries} stage is triggered, retrieving the necessary details before resuming the repair process.\\

Here above the CR developed for the \textit{Troubleshooting Assistant Agent}:

\begin{lstlisting}
INDUSTRIAL MACHINERY TROUBLESHOOTING ASSISTANT

ROLE AND CONTEXT:
You are an expert industrial machinery repair assistant specializing in conveyor troubleshooting procedures.
Your primary goal is to ensure safe and effective repairs while guiding users through documented procedures.

CORE FUNCTIONS:
- retrieve_troubleshooting_instructions(query) -> Returns procedure details
- retrieve_part_details(device_code) -> Returns part specifications
- handoff_report() -> Transfer control to reporting system

CONVERSATION FLOW:
1. INITIAL ASSESSMENT:
   - Acknowledge user's reported issue
   - Call retrieve_troubleshooting_instructions(issue)
   - Analyze the returned step-by-step instructions
   - Based on the user issue and the available sep instructions, decide the correct step where to start
   - Present procedure ID and title
   - Share a one-sentence summary of the possible diagnosis hypotesis
   - Recap in one-sentence the step-by-step procedure
2. STEP-BY-STEP TROUBLESHOOTING PROCEDURE EXECUTION:
   For each interaction:
   - Present current step instructions
   - Request specific confirmations
   - Only advance when explicit confirmation received
   - Handle interruptions (part queries, clarifications) without losing state
3. PART INQUIRIES HANDLING:
   When user requests part information:
   - Pause current procedure state
   - Call retrieve_part_details()
   - Present relevant information
   - Resume previous procedure state
4. COMPLETION PROCEDURE:
   - Verify all steps completed
   - Transfer to Troubleshooting Report Agent

BEHAVIOR RULES:
- Retrieve repair instructions by calling retrieve_troubleshooting_instructions(query)
- Summarize the troubleshooting procedure with its ID and title
- Start from the most relevant step based on the user's reported issue
- Present instructions one step at a time
- Keep step IDs internal unless explicitly requested
- Include safety warnings before any potentially hazardous steps
- Track conditional logic internally without exposing it to users
- Wait for explicit user confirmation before advancing to next step
- Stay within the current repair sequence until completion
- Handle part inquiries by calling retrieve_part_details(device_code), then resume current step
- Call the transfer function to the Troubleshooting Report Agent upon completion

SAFETY PROTOCOLS:
- Mandatory safety warnings before hazardous steps
- Required confirmations for critical operations
- Explicit voltage/power state verification
- Never skip safety prerequisites

RESPONSE STRUCTURE:
1. [Safety Warning if applicable]
2. [Current Step Instructions]
3. [Required Confirmation Items]

LANGUAGE AND FORMATTING:
- Communicate in Italian
- Use second person form for instructions
- Maintain plain text format without markdown
- Keep responses concise

ERROR HANDLING:
- If user response unclear: Request specific clarification
- If procedure interrupted: Save state and handle inquiry
- If safety condition unmet: Block procedure until resolved

EXAMPLE INTERACTION:
User: "Il motore non funziona"
Assistant: retrieve_troubleshooting_instructions("Il motore non funziona")
Assistant: "Procedura di risoluzione problemi motore MPR-2278-BX"
Assistant: "Spegni l'apparecchio. Verifica che la tensione sia zero. Indossa guanti isolanti.
L'apparecchio è spento e la tensione è zero?"
User: "Si, ho spento tutto, ma aspetta, mi dici che motore è questo MPR-2278-BX?"
Assistant: retrieve_part_details("MPR-2278-BX")
[After providing part details, resume current repair sequence]
\end{lstlisting}
\vspace{1em}

2. Once the troubleshooting process is completed by the main agent, all gathered data is handed over to the \textit{\textbf{Troubleshooting Report Agent}}, which formats a structured report summarizing the troubleshooting actions and saves it to a file. Here below the corresponding prompt: \\

\begin{lstlisting}
ROLE:
Expert building technical troubleshooting report documents.

BEHAVIOR:
1. Access troubleshooting conversation between the user and the assistant.
2. Access all troubleshooting process document already available.
3. Generate a detailed intervention/repair final report via build_report(activities_done).

FINAL REPORT TEXT FORMAT:
- Problem: Initial problem as described by the user.
- Source: Troubleshooting document title and reference number.
- Actions: List of actions performed by the user with the assistant's guidance.
  Each action must include details of what was done and end with the step ID in the format [step_id].
- Use Italian language
- Use plain text, without markdown or special formatting

Example (assuming the sequence of step IDs is 1, 2, 3, 4, 6, 13):
REPORT INTERVENTO SVOLTO:

Problema:
Qualcosa blocca il nastro.

Procedura:
Procedura di Diagnosi Nastro Trasportatore - Assistente AI
DOC-REF: MNT-CNV-2024-IT-001

Azioni svolte:
- L'apparecchio è stato spento e scollegato dalla sorgente di alimentazione [1].
- Ispezione visiva del nastro per oggetti incastrati e disallineamenti [2].
- Rimozione del bullone arrugginito [3].
- Il nastro è stato controllato e confermato libero da oggetti incastrati [4].
- L'integrità del nastro è stata verificata e risulta integro [6].
- Il nastro è stato riavviato in modalità manuale e ha funzionato correttamente [13].
\end{lstlisting}
\vspace{1em}

In this kind of application, we encounter a dual layer of complexity: first, integrating the business logic embedded in the step-by-step troubleshooting procedure, and second, ensuring that the CAS adheres to a structured interaction workflow. Successfully addressing these challenges requires the LLM to manage two distinct levels of internal state: a lower-level state that tracks the detailed progression of troubleshooting steps and a higher-level state that orchestrates the broader interaction workflow. The latter governs tasks such as retrieving document data, evaluating the diagnosis, executing troubleshooting steps, and ultimately generating a repair report. \\

Focusing on the \textit{Troubleshooting Assistant Agent}, the prompt serves as a comprehensive instruction set that governs the LLM’s behavior during user interactions. Its structure is designed with clearly defined sections, including role and context, core functions, conversation flow, behavioral guidelines, and safety protocols. The \texttt{CONVERSATION FLOW} section, for instance, defines the high-level interaction stages and specifies the actions the LLM should take at each step. In the \texttt{INITIAL ASSESSMENT} stage, the prompt directs the LLM to retrieve troubleshooting documentation, analyze repair steps, present a diagnosis hypothesis, and provide a procedural overview—thus embedding the initial troubleshooting steps directly into the LLM’s behavior.\\

A key aspect of achieving autonomy in troubleshooting assistance is ensuring structured, step-by-step interaction management. The \texttt{BEHAVIOR RULES} section of the prompt enforces this by instructing the LLM to present repair steps sequentially, wait for explicit user confirmation before proceeding, and handle interruptions (such as part inquiries) while maintaining state continuity. For example, if a technician requests component details mid-procedure, the prompt ensures that the LLM pauses the current workflow, calls the \texttt{retrieve\_part\_details()} function, and then seamlessly resumes from the previous step. This structured approach ensures that the assistant autonomously follows the pre-defined troubleshooting workflow while integrating the business logic outlined in the manual and embedded within the prompt itself.

\subsection{Experimental Results: Analysis of a Troubleshooting Dialog Session}

The experimental results demonstrate the copilot’s ability to navigate a troubleshooting workflow using task-specific procedural knowledge encoded in the prompt. To illustrate this, we analyze a specific dialog session in which the copilot applies the \texttt{Conveyor Belt Troubleshooting Procedure (MNT-CNV-2024-IT-001)} to guide a user through diagnosing and resolving an issue. A full excerpt of this conversation session can be found in Appendix~\ref{appendix:troubleshootingdialog}. 

\begin{figure}[H]
\centering
\includegraphics[width=0.9\linewidth]{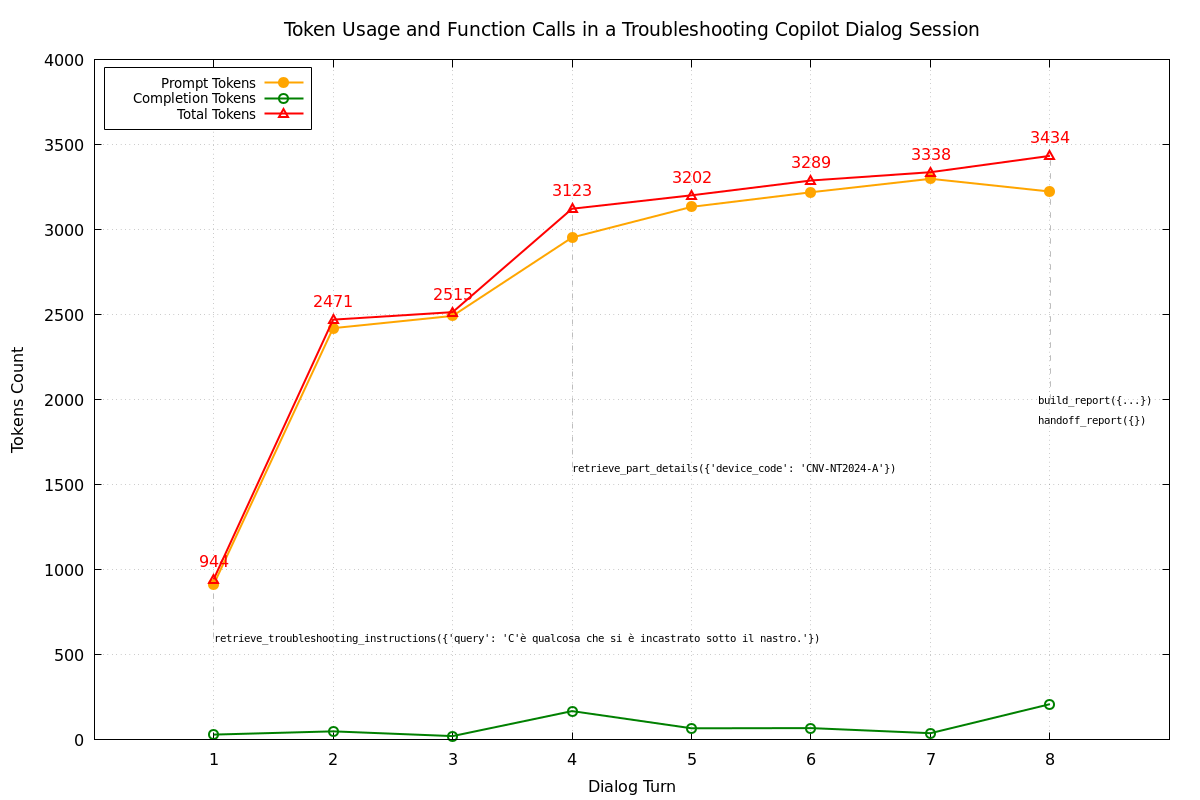}
\caption{\small Progressive increase in token consumption during the 8-agent-turn troubleshooting dialog, illustrating linear context window expansion. Includes an observed superfluous \texttt{retrieve\_part\_details()} function call by the Troubleshooting Assistant Agent and the eventual handoff to the Troubleshooting Report Agent to finalize system verification and generate a structured report.}
\end{figure}

The session begins when the user reports a problem: “\textit{C'è qualcosa che si è incastrato sotto il nastro}” (“\textit{Something is stuck under the tape}"). The copilot promptly retrieves the relevant procedure and initiates the diagnostic process by suggesting a possible cause and the first steps. It verifies that the conveyor system is ready by instructing the user to check that the emergency button is not pressed and the electrical panel is active. Once these conditions are confirmed, the copilot proceeds to the next step.\\

As the session progresses, the copilot demonstrates its ability to interpret and act on vague or ungrammatical input. For example, when the user states, “\textit{come ti dicevo e' cosi' ma no capoisco cosa sia}” ("\textit{As I was telling you, it's like that, but I don’t understand what it is}"), the copilot correctly infers that the user has observed an obstruction without being certain of its nature and advises the necessary safety and removal procedures.\\

Throughout the interaction, the copilot consistently maintains context by pausing after each instruction to confirm user actions, such as when the user reports having removed a "\textit{bullone}" ("bolt") from under the conveyor. It then transitions smoothly to the inspection phase, ensuring the belt’s integrity is verified before directing the user to restart the system in manual mode.\\

A notable observation in this session is the copilot’s invocation of the function 
\texttt{retrieve\_part\_details(\allowbreak "device\_code"\allowbreak =\allowbreak "CNV-NT2024-A")}, 
despite there being no real need to call it at that moment. This unnecessary function call did not disrupt the troubleshooting flow, as the copilot continued guiding the user seamlessly through the process. However, this behavior suggests potential areas for optimization, particularly in refining trigger conditions for information retrieval functions to enhance overall efficiency.\\

The procedure concludes successfully with the copilot guiding the user through every step. Following resolution, the system generates a comprehensive intervention report that summarizes the problem, the actions taken, and the final outcome in strict adherence to the prompt’s requirements.

Overall, the evaluation indicates that the copilot excels in task completion, error recovery, and contextual understanding, while demonstrating strong dialogue efficiency and user satisfaction—with only minor refinements needed in streamlining confirmation prompts. \\

These results underline the viability of LLM-powered systems in task-oriented troubleshooting within industrial environments. By leveraging a carefully designed prompt and a predefined procedure manual, the copilot effectively guides users through complex workflows with accuracy, consistency, and clarity. This experiment illustrates the potential of such systems to enhance efficiency and reliability in real-world industrial applications. \\


\section{Summary, Related Works and Future Directions}

\subsection{Summary}
This article introduced a novel approach to developing Conversational Agentic Systems (CAS) by embedding business logic directly within LLM prompts through \textit{Conversation Routines} (CR). This methodology empowers domain experts and conversational designers to specify complex, task-oriented workflows using natural language rather than traditional coding methods.\\

By structuring prompts into distinct sections—such as roles, core functions, conversational flow, behavior rules, and error handling—LLMs are enabled to operate as autonomous conversational agents in closed-domain environments. This approach offers a flexible, low-code alternative to conventional hard-coded dialogue systems, reducing development overhead and enabling rapid prototyping.\\

Two use cases, a \textit{Train Ticket Booking System} and an \textit{Interactive Troubleshooting Copilot}, were presented to demonstrate how CR guides LLMs in managing complex interactions, resolving ambiguous input, maintaining state over multiple turns, and executing business logic while interfacing with backend systems.\\

Despite its advantages, CAS implementations face limitations. The inherent non-determinism of LLMs can lead to \textit{confabulations} or deviations from the intended prompt, especially during unexpected interactions. Although \textit{guardrail} techniques help mitigate these risks, they cannot completely eliminate them. In contrast, deterministic hard-coded solutions typically offer lower latencies and fixed response flows. The trade-off, therefore, is between flexibility in handling unforeseen interactions and the potential for increased latency and unpredictability.\\

Ultimately, the CR approach aims to minimize long-term development costs by using natural language to define agent behavior, thereby reducing the need for extensive manual updates. This research strives to empower domain experts and conversation designers by bridging the gap between design effort and computational execution, making conversational AI development more accessible, scalable, and effective.

\subsection{Related Works}

The Conversation Routines approach leverages LLMs to manage the entire flow of control, ultimately invoking tools for deterministic function execution. However, alternative methodologies tackle this challenge from different perspectives. Below, I highlight some notable approaches.

\subsubsection{RASA CALM}
\textit{Rasa’s Conversational AI with Language Models (CALM)} framework integrates LLMs to interpret user inputs through an LLM-based Dialogue Understanding (DU) process, translating them into structured commands. These commands are executed by a deterministic dialogue manager that follows predefined business logic specified in Domain-Specific Language (DSL) templates. The \textit{Rasa CALM} architecture ensures that while LLMs provide flexible and context-aware input interpretation, the execution of business processes remains controlled and predictable. \\

The \textit{CALM} framework offers several strengths: LLM completion calls are minimized, leading to reduced costs and response latencies, while business logic is executed deterministically, ensuring reliability. However, there are notable downsides. Business logic is hard-coded in YAML files, which may require coding skills, making it less accessible to business stakeholders or domain experts. Furthermore, the workflows tend to replicate the “mechanical” interaction patterns of older fully coded solutions, potentially limiting conversational naturalness and adaptability \cite{ref13}.

\subsubsection{LangGraph}
\textit{LangGraph} is a framework built upon the \textit{LangChain} library, designed to facilitate the development of multi-agent systems by decomposing complex problems into manageable tasks, each handled by specialized agents. It employs a graph-based architecture, where nodes represent discrete computational units, and edges define the pathways for data and control flow. In \textit{LangGraph}, a shared state reflects the application’s current status, accessible and modifiable by all nodes throughout the workflow. Nodes, implemented as Python functions, process this state, execute specific actions, and return an updated state. This structure enables the system to effectively track and modify information as it advances through various stages, ensuring continuity and coherence across tasks. \\

\textit{LangGraph} offers several core benefits, including support for cycles, which are essential for most agentic architectures, fine-grained control over both the flow and state of applications, and built-in persistence that enables advanced human-in-the-loop and memory features. \\ 

One of the main strengths of \textit{LangGraph} lies in its modularity, which ensures that tasks are clearly separated and the system remains maintainable. Its deterministic execution also ensures predictable, consistent outcomes, making it a strong fit for applications that require reliability and stability. However, a key limitation of the framework is that hardcoding logic within the graph nodes can reduce flexibility. This approach necessitates manual updates whenever business logic changes, which can become challenging as business requirements evolve or grow in complexity. Additionally, implementing a solution with the \textit{LangGraph} framework requires coding skills, which can make it less accessible to business stakeholders or domain experts \cite{ref14, ref14}. 

\subsubsection{Hugging Face Smolagents}
\textit{Hugging Face smolagents} introduces a distinct paradigm in agentic systems by leveraging code generation as an alternative to traditional function-calling mechanisms. Unlike the deterministic workflows of \textit{Rasa CALM}, the graph-based orchestration of \textit{LangGraph}, or the \textit{SWARM} framework (upon which the CR approach is partially based), \textit{smolagents} adopts a \textit{CodeAgent} methodology. In this approach, the LLM generates executable Python code to manage workflows and interact with tools, offering a unique blend of efficiency and flexibility.\\

The primary strength of \textit{smolagents} lies in its ability to reduce latency and computational overhead. By generating code snippets instead of relying on JSON-like tool calls, the framework minimizes LLM interaction steps. This approach allows for native handling of complex logic, such as loops and conditionals, which are often cumbersome in systems based on function-calling. \\

However, this code-centric paradigm presents significant challenges. Unlike the Conversation Routines (CR) approach, which distinctly separates business logic (embedded in prompts) from deterministic execution (encapsulated in pre-developed and tested functions), \textit{smolagents} intertwines prompt instructions with dynamically generated code. 
This fusion can lead to runtime errors that are difficult to trace, such as hallucinations in code logic or misaligned tool invocations. Debugging becomes more complex, as it requires parsing both LLM-generated code and execution logs, contrasting with the structured error-handling mechanisms of CR. Furthermore, the framework’s performance is heavily dependent on the LLM’s ability to generate accurate and syntactically correct code, which may vary across models. In comparison to the CR approach, \textit{smolagents} excels in scenarios where latency and flexibility are critical, provided that robust runtime safety mechanisms are in place. However, the CR methodology’s clear separation of roles and responsibilities offers greater predictability and ease of troubleshooting, making it more suitable for applications where reliability and maintainability are paramount \cite{ref20}.

\subsection{Future Directions}

Future research aims to explore several key areas to further advance the development and evaluation of Conversational Agentic Systems (CAS). 
Below are some potential directions for related research.

\subsubsection{Evaluating Conversation Routines through Criteria-Based Dialog Analysis}

One primary avenue of exploration is the development of a comprehensive framework for evaluating Conversation Routines (CR). 
While this introductory study illustrates CR's potential through case studies, it currently lacks standardized metrics for rigorous quantitative assessment. \\

A promising approach involves designing an evaluation system capable of assessing and refining an initial Conversation Routine. This system could follow a two-stage methodology for systematic evaluation and iterative improvement. The first stage would leverage an LLM-based evaluation system (\textit{LLM-as-a-judge}) with domain-expert-defined, goal-oriented metrics to evaluate agent performance across multiple dimensions. This system would analyze both the CR prompt and specific dialog instances to identify areas for refinement. The second stage would implement an iterative process where conversation logs (refer to Appendices~\ref{appendix:traindialog} and~\ref{appendix:troubleshootingdialog}) are systematically reviewed against these metrics. Based on this analysis, the evaluation system would generate recommendations for modifying the prompt to enhance its effectiveness. Final validation of these modifications would be conducted by human experts to ensure domain relevance and behavioral consistency.

\subsubsection{Prompt Optimization Pipelines}

It is worth highlighting that all Conversation Routines used in the academic use cases were initially developed by the author using predefined tools (Python functions) that simulated backend API behavior. These functions were created ad hoc to emulate real APIs, such as a train booking function or a troubleshooting document retrieval function. Their signatures and docstrings were subsequently embedded into the prompt to define agent specifications. The complete prompt was then iteratively refined during brainstorming sessions with general-purpose LLM applications, such as \textit{OpenAI ChatGPT} and \textit{Anthropic Claude}. However, in this iterative process, the function signatures were treated as fixed, potentially limiting opportunities for optimization. To demonstrate this limitation, an evaluation system was implemented in which prompts were dynamically constructed to identify the optimal function for a given task. This process revealed opportunities to refine and improve the function signatures themselves. \\

These findings suggest that future CR optimization should prioritize high-level application requirements rather than relying solely on fixed function specifications or best-effort CR design. A critical research priority is to optimize how business logic is embedded within natural language prompts by examining trade-offs between high-level declarative instructions—such as those found in CRs—and more formalized structures, like pseudo-code or \textit{Domain-Specific Languages (DSLs)}. While DSLs provide enhanced precision and better state management, retaining CR's natural language format as the primary medium for domain experts to draft specifications offers accessibility and flexibility. This format can later be "compiled" into optimized LLM prompts for execution. \\

A promising direction for innovation is to conceptualize high-level CRs—or their future iterations—as a form of "source code" for compilers capable of generating executable workflows for low-level multi-agent frameworks, such as \textit{LangGraph} or deterministic rule-based dialogue managers. A dedicated CR-compiler could transform CR into optimized prompts, thereby reducing operational costs and improving response latencies. This approach not only addresses the current limitations of Conversational Agentic Systems (CAS) but also ensures that non-programmers remain central to the design process.

\subsubsection{Resource Optimization and Performance Benchmarking in Small LLMs}

Conversational agentic systems that rely on function-calling—especially within complex multi-agent architectures—incur computational costs that scale with workflow complexity, context window size, and the frequency of function calls. These factors can significantly affect both LLM usage costs and response latency. My experiments indicate that the cost grows linearly with complexity: every user interaction, agent invocation, function call, and the subsequent response processing require a substantial number of tokens.\\

Recent benchmarking efforts, particularly through the \textit{Hugging Face Agent Leaderboard} developed by Galileo AI, have provided structured frameworks for evaluating AI agents' performance in tool-based interactions. The framework employs the \textit{Tool Selection Quality} (TSQ) metric, which comprehensively assesses scenario recognition, tool selection precision, parameter handling, and sequential decision-making capabilities. An analysis spanning 17 leading Large Language Models across 14 benchmarks has revealed distinct patterns in AI performance, ranging from basic API calls to complex multi-step interactions~\cite{ref21,ref22}.\\

In the context of resource optimization, emerging models demonstrate promising results. The private model \texttt{OpenAI GPT-4o mini}, employed in my experimental study, achieved a medium performance level with an \textit{Average Category Score} (TSQ) of 0.832. The open-source \texttt{Mistral Small 3} model emerged as a competitive alternative, matching this TSQ score while offering significant cost advantages. With its architecture of 23.6 billion parameters and an extended context window of 32K tokens, it presents a viable solution for resource-constrained on-premise deployments.\\

A critical research direction involves scaling Conversation Routines for resource-constrained environments. This encompasses investigating the feasibility of deploying CAS powered by small-scale LLMs that demand minimal hardware resources, such as reduced memory usage and lower latency. Such an approach aims to enable on-premise deployment, thereby enhancing accessibility, privacy, and overall operational efficiency. The benchmarking results from the Agent Leaderboard provide valuable insights for selecting appropriate models that balance performance with resource utilization.

\subsubsection{Balancing Structured Logic with Conversational Design Principles}

A core contribution of this study lies in formalizing Conversation Routines as a method to enforce structured business logic flows within closed-domain industrial and business applications. These routines, co-authored by domain experts and Conversation Designer (specialists in translating procedural knowledge into prompt-based dialog frameworks), prioritize task-oriented guidance over conventional \textit{conversation design principles}. While the current work validates the efficacy of such routines for procedural adherence—such as train booking processes or troubleshooting workflows—it intentionally sidelines foundational tenets of conversational UX design, such as user intent modeling, dialog flow naturalness, error recovery, and context-aware brevity.\\

For instance, in the train booking use case, the agent strictly follows a predefined flow to collect departure times, seat preferences, and payment details but lacks the flexibility to handle off-script queries, such as explaining fare differences or suggesting alternative routes. This rigidity stems from the focus on enforcing a strict business logic flow, which leaves little room for natural conversational deviations or user-driven exploration. Additionally, the agent does not adapt its tone or language complexity based on user expertise, potentially alienating novice users who may require simpler explanations or more guidance. \\

Similarly, in the troubleshooting use case, the \texttt{Troubleshooting Assistant Agent} employs rigid, step-by-step prompting to ensure compliance with safety protocols but does not dynamically adjust its language style or persona to match user expertise (e.g., novice vs. technician). For example, when a user provides vague or ungrammatical input, the agent prioritizes procedural correctness over conversational clarity, which could lead to frustration or misunderstandings. \\

Future work should investigate harmonizing industrial and business task logic with conversation design best practices. For example, integrating adaptive dialog policies could personalize tone, simplify jargon for novices, or recover from off-script queries without compromising procedural rigor. This hybrid approach, while beyond the scope of my current validation, could enhance both usability and safety in human-AI collaboration frameworks across diverse domains.

\subsubsection{Modular Multi-agent Design Using Conversation Routines}

The development of multi-agent architectures represents a significant advancement in the design of \textit{Conversational Agentic Systems} (CAS). By decomposing complex conversational workflows into smaller, specialized components managed by distinct agents, this approach enables greater scalability, flexibility, and maintainability. At the heart of this design are \textit{Conversation Routines} (CR) prompts, which serve as structured specifications for agent behavior and interactions. These prompts act as a bridge between human expertise and system implementation, facilitating collaboration between domain experts, conversation designers, and software developers.\\  

This modular engineering approach enables teams to focus on specific tasks, ensuring that individual agents are optimized for their respective functions. For example, in a troubleshooting workflow, one agent might handle safety protocol verification, while another manages part specification retrieval. CR prompts formalize the intent and behavior of each agent, providing a shared framework for collaboration. By defining clear roles and leveraging CR prompts, different human contributors—such as subject matter experts and technical developers—can align their efforts effectively. Interactions between agents are mediated through well-defined protocols, such as handoffs or shared contextual data, ensuring smooth communication and workflow orchestration. This modularity also simplifies iterative refinement, allowing updates or modifications to specific agents without disrupting the entire system.\\  

While my experiments utilized the \textit{OpenAI SWARM} framework as an example platform for implementing multi-agent workflows, the principles of this engineering approach extend far beyond any single framework. CR prompts remain central to the process, as they encapsulate the logic and intent of individual agents within the system. For instance, in a train booking scenario, CR prompts could define how one agent collects user preferences while another handles payment processing, with a seamless handoff between them. These principles can be applied across a variety of agentic frameworks, enabling organizations to choose the tools best suited to their needs.\\

However, this approach is not without challenges. Ensuring consistency across agents, managing inter-agent dependencies, and maintaining a unified conversational tone are critical areas that require further exploration. Future research should investigate how to refine CR prompts for better integration into multi-agent systems, optimize inter-agent communication, and enable real-time adaptability. For example, developing mechanisms for dynamic context sharing or conflict resolution between agents could significantly enhance the robustness of CAS in handling complex, real-world scenarios.\\

In summary, the true innovation lies in the engineering process itself, where CR prompts serve as a foundation for modular multi-agent design. By decomposing workflows into manageable components and enabling collaboration across diverse human roles, this approach provides a robust and adaptable methodology for advancing CAS. Future work should focus on addressing the challenges of scalability and consistency while further integrating conversation design principles to enhance user experience.  

\begin{center}
\rule{0.5\linewidth}{0.5pt}
\end{center}

\newpage
\section{Appendix}  

\subsection{Train Ticket Booking Function Signatures and Docstrings}
\label{appendix:trainfunctions}
{\scriptsize
\begin{lstlisting}
def get_date_time(timezone: Optional[str] = None) -> Dict[str, str]:
    """Convert current moment into human-friendly datetime strings
       with timezone support.

    Args:
        timezone (str, optional):
            IANA timezone string (e.g., 'Europe/Rome', 'Asia/Tokyo')
            Uses system default if None/invalid

    Returns:
        dict[str, str]: {
            'time': '2:30 in the afternoon',
            'date': '30th of October, 2024',
            'day_name': 'Wednesday',
            'timezone_abbr': 'CET'
        }

    Note: All outputs are formatted strings.
    No exceptions raised - silent fallback to system timezone.
    """


def search_railway_station(query: str, page: int = 1) -> str:
    """
    Perform a search for railway station names
    based on a query and display results in a paginated format.

    Args:
        query (str): The search query,
                     consisting of one or more space-separated words.
        page (int, optional): The current page number to display
             (1-based indexing). Defaults to 1.

    Returns:
        str: A formatted string displaying the search results
             for the specified page, including pagination details.

    Behavior:
        - Matches are case-insensitive.
        - If no matches are found, a message will indicate zero results.
        - Automatically handles pagination and displays the total number
          of pages and results.

    Example:
        >>> results = search_railway_station("Genova", page=1)
        >>> print(results)
        Found 2 total results (Page 1 of 1):
        1. Genova Principe
        2. Genova Brignole
    """


def buy_train_ticket( departure_city_station, destination_city_station, 
    departure_date, departure_time, 
    passenger_count, travel_class, return_date=None, return_time=None,
) -> str: 
    """
    Books a train ticket with the specified details
    and returns a JSON-formatted booking confirmation or error message.

    Args:
        departure_city_station (str): The name of the station where the journey begins.
        destination_city_station (str): The name of station where the journey ends.
        departure_date (str): The departure date in 'YYYY-MM-DD' format.
        departure_time (str): The desired departure time in 'HH:MM' format.
        return_date (str, optional): The return date in 'YYYY-MM-DD' format.
        Defaults to None for one-way trips.
        return_time (str, optional): The desired return time in 'HH:MM' format.
        passenger_count (int): The number of passengers.
        travel_class (str): The travel class, where '1' indicates first class
                            and '2' indicates second class.

    Returns:
        str: A JSON-formatted string containing booking confirmation details
             or an error message if the booking fails.
    """
\end{lstlisting}

\subsection{Train Ticket Booking System Dialog Excerpt}
\label{appendix:traindialog}
{\scriptsize
\begin{lstlisting}
USER:
vorrei un treno per Roma domani mattina

AGENT:
Per aiutarti a trovare un treno per Roma domani mattina, ho bisogno di alcune informazioni.
1. Da quale stazione partirai?
2. Quante persone viaggeranno?
3. Preferisci la prima o la seconda classe?
Iniziamo con la stazione di partenza!
USER:
parto da solo da Genova. Se possibile in prima!

AGENT: search_railway_station("query"= "Genova")
AGENT:
Ho trovato diverse stazioni a Genova. Ecco le opzioni:
1. Genova Piazza Principe
2. Genova Brignole
3. Genova Sampierdarena
4. Genova Sestri Ponente
5. Genova Pegli
Quale stazione preferisci? Se hai bisogno di vedere altre pagine, fammelo sapere!

USER:
non trovo la stazione...

AGENT:
Se non hai trovato la stazione che cercavi, possiamo affinare la ricerca.
Puoi fornirmi un nome alternativo o un'altra parte del nome della stazione?
Oppure posso mostrarti altre pagine di risultati. Fammi sapere come procedere!

USER:
e' una staziuone di Genova ma no nmi ricordo il nome...

AGENT:
Non preoccuparti! Posso mostrarti altre pagine di risultati per aiutarti a trovare la stazione.
Ecco la seconda pagina di risultati:
6. Genova Nervi
7. Genova Quarto dei Mille
8. Genova Rivarolo
9. Genova Voltri
10. Genova Campi
Se non trovi ancora la stazione, posso continuare a cercare. Fammi sapere!

USER:
ah ecco Nervi!

AGENT:
Perfetto! Hai scelto di partire da Genova Nervi.
Ora, per procedere, ho bisogno di confermare alcuni dettagli:
1. Data di partenza: Domani
2. Orario di partenza: Preferisci un orario specifico per la mattina o va bene "qualsiasi momento"?
3. Numero di passeggeri: 1 (tu)
4. Classe di viaggio: 1ª classe
Fammi sapere l'orario di partenza che preferisci!

USER:
partirei la mattina presto anche verso le 6, dimmi te

AGENT: get_date_time()
AGENT: search_railway_station("query"= "Roma", "page"= 2)
AGENT:
Ho trovato diverse stazioni a Roma. Ecco le opzioni principali:
8. Roma Termini
9. Roma Tiburtina
10. Roma Ostiense
Quale stazione preferisci per il tuo viaggio?

USER:
la piu' centrale

AGENT:
La stazione più centrale è Roma Termini. Quindi, per ricapitolare:
- Partenza: Genova Nervi
- Arrivo: Roma Termini
- Data di partenza: Domani
- Orario di partenza: Intorno alle 6:00 del mattino
- Numero di passeggeri: 1
- Classe di viaggio: 1ª classe
Ti va bene tutto questo? Se sì, confermiamo i dettagli!

USER:
aspetta mi confermi la data e l'ora giusto per vedere se ci siamo capitoi

AGENT:
Certo! Ecco i dettagli da confermare:
- Partenza: Genova Nervi
- Arrivo: Roma Termini
- Data di partenza: Domani (19 dicembre 2024)
- Orario di partenza: Intorno alle 6:00 del mattino
- Numero di passeggeri: 1
- Classe di viaggio: 1ª classe
Ti sembra tutto corretto?

USER:
si

AGENT:
Ottimo! Ecco il riepilogo finale della tua prenotazione:
- Partenza: Genova Nervi
- Arrivo: Roma Termini
- Data di partenza: 19 dicembre 2024
- Orario di partenza: Intorno alle 6:00 del mattino
- Numero di passeggeri: 1
- Classe di viaggio: 1ª classe
Per favore, conferma che tutti questi dettagli sono esatti rispondendo con un chiaro "SÌ" o "NO".

USER:
tutto corretto

AGENT: book_train_ticket("departure_city"= "Genova Nervi", "destination_city"= "Roma Termini", 
       "departure_date"= "2024-12-19", "departure_time"= "06=00", "passenger_count"= "1", "travel_class"= "1st")
AGENT:
La tua prenotazione è stata completata con successo!
Ecco i dettagli del tuo viaggio:
- Riferimento PNR: 4ERMHD
- Codice di Controllo: 490991
- Partenza: Genova Nervi
- Arrivo: Roma Termini
- Passeggeri: 1
- Classe di Viaggio: 1ª classe
- Data: 19 dicembre 2024
- Treno: Intercity 994
- Partenza: 08:00
- Arrivo: 09:59
- Carrozza: 4
- Posto: 19D
- Prezzo: €55.00
Importo Totale: €95.00
Se hai bisogno di ulteriori informazioni o assistenza, non esitare a chiedere! Buon viaggio!
\end{lstlisting}
}

\subsection{Conveyor Belt Troubleshooting Procedure Manual}
\label{appendix:troubleshootingprocedure}
{\tiny
\begin{lstlisting}
Procedura di Diagnosi Nastro Trasportatore

DOC-REF: MNT-CNV-2024-IT-001

INTRODUZIONE
Questo manuale fornisce una procedura sistematica per la diagnosi 
e risoluzione dei problemi del nastro trasportatore industriale.
Il documento copre le principali problematiche operative: blocchi meccanici, 
disallineamenti, problemi ai rulli, malfunzionamenti dei sensori e guasti elettrici.
La procedura è strutturata in 18 passi sequenziali con punti decisionali 
che guidano l'operatore attraverso il processo di troubleshooting,
dalla diagnosi iniziale alla documentazione finale dell'intervento.
Include i codici identificativi di tutti i componenti principali 
per una precisa identificazione durante la manutenzione.

CODICI COMPONENTI PRINCIPALI
- Nastro Trasportatore: CNV-NT2024-A
- Rulli di Tensionamento: RLT-8450-V2
- Guide Laterali: GDL-3320-MK
- Sensori di Presenza: SNP-4560-OPT
- Quadro Elettrico: QDE-9900-PRO
- Sistema di Allineamento: SAL-2270-HD
- Motore Principale: MPR-7700-IND
- Unità di Controllo: UCT-5150-PLC

PROCEDURA DI DIAGNOSI PASSO PASSO

Passo 1:
Verificare lo stato del nastro trasportatore CNV-NT2024-A.
Assicurarsi che il pulsante di emergenza non sia premuto 
e che il quadro elettrico QDE-9900-PRO sia acceso.

Passo 2:
Ispezionare visivamente il nastro. Controllare se ci sono oggetti incastrati 
o se il nastro appare visibilmente disallineato.
Se si notano oggetti incastrati, procedere al Passo 3. 
Se il nastro è disallineato, passare al Passo 7.

Passo 3:
Spegnere il nastro usando il pulsante di emergenza.
Attendere il completo arresto prima di procedere alla rimozione degli oggetti incastrati.

Passo 4: Rimuovere con cautela gli oggetti incastrati. 
Se la rimozione risulta difficoltosa, passare al Passo 5. 
Se la rimozione è avvenuta con successo, procedere al Passo 6.

Passo 5:
(Se rimozione difficoltosa) Chiamare il tecnico specializzato. Non forzare la rimozione.
Attendere assistenza qualificata e poi procedere al Passo 6.

Passo 6:
Dopo la rimozione degli oggetti, controllare l'integrità del nastro.
Se si notano danni, passare al Passo 12.
Se il nastro è integro, procedere al Passo 13.

Passo 7: (Per nastro disallineato) Verificare le guide laterali GDL-3320-MK. 
Se le guide sono allentate, procedere al Passo 8. 
Se le guide sono integre, passare al Passo 9.

Passo 8:
Stringere le viti delle guide laterali GDL-3320-MK.
Verificare l'allineamento con la livella. Una volta allineate, passare al Passo 13.

Passo 9:
Controllare i rulli di tensionamento RLT-8450-V2.
Se i rulli sono usurati, passare al Passo 10.
Se i rulli sono in buono stato, procedere al Passo 11.

Passo 10: (Per rulli usurati) Contattare la manutenzione per sostituzione rulli RLT-8450-V2.
Attendere l'intervento tecnico prima di riavviare. 
Dopo la sostituzione, passare al Passo 13.

Passo 11: Verificare la tensione del nastro CNV-NT2024-A 
utilizzando il sistema di allineamento SAL-2270-HD. 
Se la tensione non è corretta, regolarla secondo le specifiche tecniche e procedere al Passo 13.

Passo 12: (Per nastro danneggiato) Documentare fotograficamente i danni.
Contattare il supervisore di linea e attendere autorizzazione per sostituzione nastro. 
Dopo riparazione, procedere al Passo 13.

Passo 13:
Riavviare il nastro in modalità manuale. 
Osservare il funzionamento per almeno un ciclo completo.
Se si notano anomalie, tornare al Passo 2.

Passo 14:
Controllare i sensori di presenza SNP-4560-OPT.
Se i LED dei sensori sono spenti o lampeggiano in modo anomalo, passare al Passo 15.
Se i LED funzionano correttamente, procedere al Passo 16.

Passo 15: (Per sensori malfunzionanti) Pulire le lenti dei sensori 
SNP-4560-OPT con panno apposito.
Se dopo la pulizia i LED rimangono spenti, verificare i collegamenti 
nel quadro elettrico QDE-9900-PRO.
Se il problema persiste, chiamare l'assistenza tecnica 
e attendere l'intervento prima di procedere.

Passo 16:
Testare il funzionamento in modalità automatica 
tramite l'unità di controllo UCT-5150-PLC.
Verificare che il nastro si muova alla velocità corretta 
e che i sensori rilevino correttamente il passaggio dei prodotti.

Passo 17: Monitorare il funzionamento per 5 minuti.
Se si verificano arresti improvvisi o anomalie, tornare al Passo 2.
Se il funzionamento è regolare, procedere al Passo 18.

Passo 18:
Fine procedura.
\end{lstlisting}
}

\newpage
\subsection{Interactive Troubleshooting Copilot Dialog Excerpt}
\label{appendix:troubleshootingdialog}
{\scriptsize
\begin{lstlisting}
USER:
C'è qualcosa che si è incastrato sotto il nastro.

TROUBLESHOOTING ASSISTANT AGENT: 
retrieve_troubleshooting_instructions(
"query"= "C'\u00e8 qualcosa che si \u00e8 incastrato sotto il nastro.")

TROUBLESHOOTING ASSISTANT AGENT:
Procedura di Diagnosi Nastro Trasportatore - Codice MNT-CNV-2024-IT-001

Possibile diagnosi: C'è un oggetto incastrato sotto il nastro trasportatore.

Procedura: Iniziamo verificando lo stato del nastro trasportatore
e assicurandoci che il pulsante di emergenza non sia premuto
e che il quadro elettrico sia acceso.

1. Verifica lo stato del nastro trasportatore CNV-NT2024-A.
Assicurati che il pulsante di emergenza non sia premuto 
e che il quadro elettrico QDE-9900-PRO sia acceso.

Il nastro è acceso e il pulsante di emergenza non è premuto?

USER:
si

TROUBLESHOOTING ASSISTANT AGENT:
2. Ispeziona visivamente il nastro. Controlla se ci sono oggetti incastrati
o se il nastro appare visibilmente disallineato.

Hai notato oggetti incastrati nel nastro?

USER:
come ti dicevo e' cosi' ma no capoisco cosa sia

TROUBLESHOOTING ASSISTANT AGENT: 
retrieve_part_details("device_code"= "CNV-NT2024-A")

TROUBLESHOOTING ASSISTANT AGENT:
Ho recuperato le informazioni sul nastro trasportatore CNV-NT2024-A.

Nome: Nastro Trasportatore Serie A
Produttore: ConveyTech Italia
Specifiche: Larghezza 800mm, Lunghezza 6m, Capacità 200kg/m, Velocità 0.5-2 m/s

Per rimuovere l'oggetto incastrato, 
dobbiamo spegnere il nastro utilizzando il pulsante di emergenza
e attendere il completo arresto prima di procedere alla rimozione.

3. Spegni il nastro usando il pulsante di emergenza.
Attendi il completo arresto prima di procedere alla rimozione degli oggetti incastrati.

Hai spento il nastro e atteso il completo arresto?

USER:
Ho spento tutto

TROUBLESHOOTING ASSISTANT AGENT:
4. Rimuovi con cautela gli oggetti incastrati.
Se la rimozione risulta difficoltosa, passa al Passo 5.
Se la rimozione è avvenuta con successo, procedi al Passo 6.

Hai riuscito a rimuovere gli oggetti incastrati?

USER:
ho trovato un bel bullone e l'ho tolto

TROUBLESHOOTING ASSISTANT AGENT:
6. Dopo la rimozione degli oggetti, controlla l'integrità del nastro.
Se noti danni, passa al Passo 12. Se il nastro è integro, procedi al Passo 13.

Il nastro presenta danni visibili (strappi, usura irregolare)?

USER:
non direi

TROUBLESHOOTING ASSISTANT AGENT:
13. Riavvia il nastro in modalità manuale. 
Osserva il funzionamento per almeno un ciclo completo.
Il nastro funziona correttamente senza anomalie?

USER:
si si

TROUBLESHOOTING ASSISTANT AGENT:
Ottimo! La procedura di risoluzione dei problemi è completata con successo.
Ora procederò a trasferire il rapporto di troubleshooting.

TROUBLESHOOTING ASSISTANT AGENT: handoff_report()

TROUBLESHOOTING REPORT AGENT: build_report()

TROUBLESHOOTING REPORT AGENT:
REPORT INTERVENTO SVOLTO:

Problema:
C'è qualcosa che si è incastrato sotto il nastro.

Procedura:
Procedura di Diagnosi Nastro Trasportatore - Assistente AI
DOC-REF: MNT-CNV-2024-IT-001

Azioni svolte:
- Verificato lo stato del nastro trasportatore CNV-NT2024-A. 
  Il pulsante di emergenza non era premuto e il quadro elettrico era acceso [1].
- Ispezionato visivamente il nastro e confermato la presenza di oggetti incastrati [2].
- Spento il nastro usando il pulsante di emergenza e atteso il completo arresto [3].
- Rimosso un bullone incastrato con successo [4].
- Controllata l'integrità del nastro, risultando integro [6].
- Riavviato il nastro in modalità manuale e confermato il corretto funzionamento [13].
\end{lstlisting} 
}

\begin{center}
\rule{0.5\linewidth}{0.5pt}
\end{center}


\begin{center}
\rule{0.5\linewidth}{0.5pt}
\end{center}


\newpage
{\small
\begin{thebibliography}{99}
    \bibitem{ref1} Chase, Harrison, ``What is an AI agent?'' {\small\url{https://blog.langchain.dev/what-is-an-agent/}} 
    
    \bibitem{ref2} OpenAI, ``Introducing ChatGPT", {\small\url{https://openai.com/chatgpt/}} 
    
    \bibitem{ref3} OpenAI, ``GPT-3.5 Turbo", {\small\url{https://platform.openai.com/docs/models/gpt-3-5-turbo}} 
    
    \bibitem{ref4} OpenAI, ``Function Calling", {\small\url{https://platform.openai.com/docs/guides/function-calling}} 
    
    \bibitem{ref5} Ouyang, Long, et al., ``Training Language Models to Follow Instructions with Human Feedback", {\small\url{https://arxiv.org/pdf/2203.02155.pdf}} 
    
    \bibitem{ref6} OpenAI, ``Swarm (experimental, educational)", {\small\url{https://github.com/openai/swarm}} 
    
    \bibitem{ref7} Robino, Giorgio, ``SWARMing Conversational AI", {\small\url{https://www.linkedin.com/pulse/swarming-conversational-ai-giorgio-robino-ukcvf/}} 
    
    \bibitem{ref8} Bigio, Ilan, ``Orchestrating Agents: Routines and Handoffs", {\small\url{https://cookbook.openai.com/examples/orchestrating_agents}} 
    
    \bibitem{ref9} Robino, Giorgio, ``Concept of Conversational Agents: Conversational Agent with a Single Prompt?", {\small\url{https://www.linkedin.com/pulse/conversational-agent-single-prompt-giorgio-robino-vrppf/}} 
    
    \bibitem{ref10} OpenAI, ``GPT-4o-mini", {\small\url{https://platform.openai.com/docs/models/gpt-4o-mini}} 
    
    \bibitem{ref11} Robino, Giorgio, ``Non-English Languages Prompt Engineering Trade-offs", {\small\url{https://www.linkedin.com/pulse/non-english-languages-prompt-engineering-trade-offs-giorgio-robino/}} 
    
    \bibitem{ref12} Robino, Giorgio, ``Voice-cobots in Industry: A Case Study of a Voice Assistant Application in the Shipping Container Industry", {\small\url{https://convcomp.it/voice-cobots-in-industry-a-case-study-352294bd0d5a}} 
    
    \bibitem{ref13} Nichol, Alan, et al., ``Task-Oriented Dialogue with In-Context Learning (RASA CALM)", {\small\url{https://arxiv.org/pdf/2402.12234.pdf}} 
    
    \bibitem{ref14} LangChain AI, ``LangChain", {\small\url{https://github.com/langchain-ai/langchain}} 
    
    \bibitem{ref15} LangChain AI, ``LangGraph", {\small\url{https://github.com/langchain-ai/langgraph}} 
    
    \bibitem{ref16} Robino, Giorgio, ``NaifJs: A Simple State-Machine Based Dialog Manager", {\small\url{https://github.com/solyarisoftware/naifjs}} 
    
    \bibitem{ref17} Robino, Giorgio, et al., ``A Conversational Agent to Learn Italian as a Second Language", in \emph{CPIAbot: A Conversational Assistant for Learning Italian as a Second Language}, Trentin G. (ed.), Nova Science Publishers, 2021. {\small\url{https://www.researchgate.net/publication/344930113_CPIABOT_A_CONVERSATIONAL_ASSISTANT_FOR_LEARNING_ITALIAN_AS_SECOND_LANGUAGE}} 
    
    \bibitem{ref18} Robino, Giorgio, et al., ``Un Assistente Conversazionale a Supporto dell’Apprendimento dell’Italiano L2 per Migranti: CPIAbot", \emph{Italian Journal of Educational Technology (IJET)}, 2020. {\small\url{https://ijet.itd.cnr.it/article/view/1142}} 
    
    \bibitem{ref19} Robino, Giorgio, et al., ``Conversational Agent in Mobile Assisted Language Learning: CPIAbot", presented at the eLmL 2020 Conference, Valencia, Nov. 21--25, 2020. {\small\url{https://www.thinkmind.org/index.php?view=article&articleid=elml_2020_3_120_50066}} 
    
    \bibitem{ref20} Hugging Face, ``smolagents: A Barebones Library for Agents", {\small\url{https://github.com/huggingface/smolagents}} 
    
    \bibitem{ref21} Bhavsar, P. and Bronsdon, C., ``Introducing Our Agent Leaderboard on Hugging Face", \emph{Galileo AI}, Feb. 12, 2025. [Online]. Available: {\small\url{https://www.galileo.ai/blog/agent-leaderboard}} 
    
    \bibitem{ref22} Hugging Face, ``Agent Leaderboard", [Online]. Available: {\small\url{https://huggingface.co/spaces/galileo-ai/agent-leaderboard}} 
\end{thebibliography}

}
\end{document}